\documentclass[twoside]{article}

\usepackage[accepted]{aistats2025}
\usepackage[round]{natbib}

\usepackage{algorithmic}
\usepackage{algorithm}

\usepackage{amsthm,amsmath,amssymb,mathtools,bm,bbm,mathrsfs}

\def\eqref#1{(\ref{#1})}

\def\vmu{{\bm{\mu}}}
\def\vtheta{{\bm{\theta}}}

\def\vf{{\bm{f}}}

\def\vp{{\bm{p}}}

\def\vr{{\bm{r}}}

\def\vw{{\bm{w}}}
\def\vx{{\bm{x}}}
\def\vy{{\bm{y}}}

\def\vbeta{{\bm{\beta}}}
\def\vomega{{\bm{\omega}}}
\def\vtheta{{\bm{\theta}}}
\def\vmu{{\bm{\mu}}}
\def\vpsi{{\bm{\psi}}}

\def\vkappa{{\bm{\kappa}}}
\def\Nrm{\mathcal{N}}
\def\Dat{\mathcal{D}}
\def\trp{^\top}

\def\mA{{\bm{A}}}

\def\mI{{\bm{I}}}

\def\mK{{\bm{K}}}

\def\mX{{\bm{X}}}

\DeclareMathAlphabet{\mathsfit}{\encodingdefault}{\sfdefault}{m}{sl}
\SetMathAlphabet{\mathsfit}{bold}{\encodingdefault}{\sfdefault}{bx}{n}

\theoremstyle{plain}

\newtheorem{definition}{Definition}

\newtheorem{lemma}{Lemma}

\def\1{\mathbbm{1}}

\usepackage{graphicx, url}
\usepackage[algo2e,linesnumbered]{algorithm2e}
\usepackage[utf8]{inputenc} 
\usepackage[T1]{fontenc}    
\usepackage{hyperref}       
\usepackage{pifont}
\usepackage{subfigure}

\usepackage{multirow}
\usepackage{makecell}
\usepackage{booktabs}       
\usepackage{amsfonts}       
\usepackage{nicefrac}       
\usepackage{microtype}      
\usepackage{color,colortbl}         
\usepackage[usenames,dvipsnames]{xcolor}
\usepackage{multicol}
\usepackage{footnote}
\usepackage{dsfont}

\usepackage{thm-restate}
\usepackage{enumitem}
\usepackage{commath}
\usepackage{comment}
\usepackage{tablefootnote}
\usepackage{lmodern}
\usepackage[T1]{fontenc}
\usepackage{pifont}
\let\oldding\ding 
\renewcommand{\ding}[2][1]{\scalebox{#1}{\oldding{#2}}} 
\usepackage{enumitem}

\begin{document}

%
\runningtitle{Bayesian Principles Improve Prompt Learning In Vision-Language Models}

%

\twocolumn[

\aistatstitle{Bayesian Principles Improve Prompt Learning \\
In Vision-Language Models}

\aistatsauthor{ Mingyu Kim\textsuperscript{*}\footnotemark[1] \And Jongwoo Ko\textsuperscript{*}\footnotemark[2] \And  Mijung Park\textsuperscript{$\dagger$}\footnotemark[1]}

\vspace{3pt} 
\aistatsaddress{ \footnotemark[1]UBC CS  \qquad \footnotemark[2]KAIST AI \\ 
\vspace{5pt} 
\texttt{\{mgyu.kim, mijung.park\}@ubc.ca} \quad \texttt{jongwoo.ko@kaist.ac.kr}}
]

\begin{abstract}
Prompt learning is a popular fine-tuning method for vision-language models due to its efficiency. It requires a small number of additional learnable parameters while significantly enhancing performance on target tasks. However, most existing methods suffer from overfitting to fine-tuning data, yielding poor generalizability. To address this, we propose a new training objective function based on a Bayesian learning principle to balance adaptability and generalizability. We derive a prior over the logits, where the mean function is parameterized by the pre-trained model, while the posterior corresponds to the fine-tuned model. This objective establishes a balance by allowing the fine-tuned model to adapt to downstream tasks while remaining close to the pre-trained model. 
%
To avoid the overfitting issues of the standard softmax function,
we adopt the one-vs-each softmax approximation along with its P\'olya-Gamma augmentation (OVE-PG).
We evaluate our method on several benchmark datasets and demonstrate that using the Bayesian principle for prompt learning is indeed a sensible choice. Code is available at the \href{https://github.com/ParkLabML/Bayesian_Principles_Improve_Prompt_Learning_In_Vision_Language_Models}{official repository}.
\end{abstract}


\section{Introduction}

Vision-Language models (VLMs), such as CLIP \citep{radford2021learning} and ALIGN \citep{jia2021scaling}, have been in the spotlight for their ability to generalize across unseen datasets, leveraging their universal knowledge. A key characteristic of these VLMs, particularly in image classification, is their capacity to classify novel classes without prior exposure by using simple class-specific text prompt \citep{radford2021learning, jia2021scaling}. 
To enhance this ability, recent approaches have explored efficient transfer learning by employing trainable continuous prompt embeddings, which are fine-tuned using a small set of data points. These methods allow VLMs to adapt to new tasks while avoiding the exhausted computation costs typically associated with full fine-tuning \citep{zhou2022learning, zhou2022conditional}. This approach extends their generalization in a modal-agnostic manner, which has been proven in the classification of natural language models and noisy label situations \citep{liu2023gpt, wu2023prompt}. 

However, one issue that arises in prompt learning is still the risk of overfitting. It degrades performance on unseen datasets, as fine-tuning often leads to the forgetting of pre-trained knowledge \citep{khattak2023self}. Initial attempts to mitigate have introduced knowledge distillation (KD) with pre-trained models \citep{khattak2023self, yao2023visual, oh2023towards}. The KD regularization aims to maintain global knowledge while incorporating task-specific information. In parallel, Bayesian approaches to prompt learning have emerged as a way to handle diverse representations \citep{lu2022prompt, chen2023plot, derakhshani2023bayesian, cho2024make}. These approaches treat multiple prompts as a population and model their distribution during fine-tuning. Each method designates its own form of regularization to account for the different representations that each prompt encodes.  

Despite many appealing approaches, a significant limitation is that previous approaches have developed in isolation without integrating two strengths. Specifically, KD approaches and Bayesian methods for prompt learning have been explored independently. As a result, individual method remains incompatible with one another, limiting their potential when used together. 

To address this gap, we propose a Bayesian principle for prompt learning in VLMs that seamlessly integrates both distributional learning and knowledge distillation. First, the proposed method employs P\'olya-gamma augmentation for composite likelihood, which helps accommodate the stochastic nature of the sigmoid function, thereby improving its distributional approximation for the softmax function. The standard softmax function is prone to overfitting specific one-hot target label information \citep{veličković2024softmaxforsharpoutofdistribution}, whereas the one-vs-each softmax approximation along with its p\'olya-gamma augmentation (OVE-PG) technique introduces advantageous noises and adjusts the logit functions using a prior mechanism during training \citep{snell2021bayesian}. Second, the proposed method aligns the KD regularization with the probabilistic nature. By representing all components, the proposed method synergizes with distributional logit function, leading to enhancing generalization. Notably, the proposed method remains fully compatible with previous approaches without requiring any modifications. 

Our theoretical and experimental findings indicate that this method effectively harnesses global information. More specifically, it achieves an effective balancing between task-specific learning and global information, resulting to prompt representations suited for adaptation and generalization. The key contribution of this paper includes following:

\begin{enumerate}
    \item We introduce a novel Bayesian principle, combining Polýa-Gamma augmentation and a one-vs-each softmax approximation, to mitigate overfitting in the prompt learning of VLMs. Our approach is simple and easily compatible with existing methods. We also demonstrate the efficacy of our method using 1D examples to provide clear intuition. 
    \textcolor{blue!80!black}{\textbf{(Section 3.1 -- 3.3)}} 
    
    \item We also observe that our probabilistic framework can naturally be combined with KD regularization between the prior and posterior distributions of logit function, which is the inner product of image and text embeddings. Additionally, we provide insightful empirical validation showing that controlling the strength of this regularization enhances the effectiveness of our approach. 
    \textcolor{blue!80!black}{\textbf{(Section 3.4)}} 
    
    \item We show that OVE-PG significantly improves generalization for prompt learning in VLMs, despite its simplicity, and may alleviate overfitting better than complex regularization terms or additional network parameters. 
    \textcolor{blue!80!black}{\textbf{(Section 5)}} 
\end{enumerate}


\section{Background }

In this section, we provide background knowledge on prompt learning. Below, we start by describing the CLIP model and various prompt learning approaches.

\subsection{Contrastive Language-Image Pretraining}

The base model we consider in this paper is the CLIP model \citep{radford2021learning},  which consists of an image encoder and a text encoder.
Given an image input denoted by $x$ and a corresponding text prompt denoted by $\vp$, the image encoder denoted by $I(\vx)$ outputs the $d$-dimensional image embedding, while the text encoder denoted by $T(\vp)$ outputs the text embedding where the size of the word embedding is the text length and embedding dimension.
Once the model is trained with the contrastive loss, the text embedding can serve as a linear classifier for novel datasets.
That is, the logit function can be defined by the inner product between the image embedding and text embedding, where the text embedding plays the role of linear classifier weight. Specifically, consider a deterministic\footnote{This is a deterministic function because once training is done, the image embedding is fixed, and the linear classifier weight is also fixed. Hence, given an arbitrary input, the output is deterministic.} function given an input $\vx$ and a label $c$, $f^{c}(x) :=I(\vx)^\top \vw^c $. 
Here, $w^c$ is a classifier weight for class $c$ and the classifier weight is simply the text embedding of the $c$-th class prompt $\vw^c = T(\vp^{c})$.  
Using this function,
the predictive probability can then be represented by 
$p(y|x) {=} \frac{\exp[f^{y}(x)]}{\sum_{c}^C \exp[f^{c}(x)]}$.


\subsection{Prompt Learning}
While the classifiers based on CLIP showed unprecedented zero-shot classification accuracy, their manual prompting for each example is a limiting factor. 
 \citet{zhou2022learning} and
 \citet{zhou2022conditional} proposed the new line of work, so-called \textit{prompt learning}, in which a small number of trainable parameters are defined and fine-tuned for target downstream tasks. 
 In Context Optimization (CoOp) \citep{{zhou2022learning}}, learning the embedding matrix $\vp_\vtheta$ with parameters $\vtheta$ was proposed. In this case, the logit function can be written as 
 $f^{c}_\vtheta(\vx) :=I(\vx)^\top T(\vp_\vtheta^{c})$. Here $I$ and $T$ are fixed. 
 On the other hand, in Conditional Prompt Learning (CoCoOp) \citep{zhou2022conditional}, both the tokens $\vp_{\theta_1}$ identical to CoOp and an additional set of residual prompts $\vr_{\theta_2}$ are learnable. This adjustment is designed to effectively tackle the domain shift by leveraging image-based features. In this approach, the logit function can be written as 
 $f^{c}_\vtheta(\vx) :=I(\vx)^\top T(\vp_{\vtheta_1}^{c} + \vr_{\vtheta_2}(\vx))$.
More recent work considers learning the distribution over the residual prompts and regularizing the prompt space to improve the prompt generalization on
unseen prompts, e.g., \citep{lu2022prompt, Derakhshani_2023_ICCV}. The logit function can take either of the forms in CoOp or CoCoOp:
 $f^{c}_\vtheta(\vx) = I(\vx)^\top T(\vp_{\vtheta_1}^{c} + \vr_{\vtheta_2}(\vx))$.

What distinguishes our work from existing work is that we consider an approximated softmax function. This softmax function is bounded by one-vs-each approximation, which involves applying the sigmoid function to each pair-wise comparison with logit values. More precisely, a sigmoid function is applied to the difference between class-wise logit values. Additionally, we utilize the P\'olya-gamma distribution to manage each sigmoid function effectively. We cover this in the following subsection.

\subsection{P\'olya-Gamma Data Augmentation}

P\'olya-Gamma (PG) data augmentation introduces an auxiliary variable per datapoint that is Pólya-Gamma distributed, such that the log-odds can be written as a mixture of Gaussians with respect to a Pólya-Gamma distribution \citep{PG13}:
\begin{align}
 \frac{(e^{\psi})^a}{(1 + e^{\psi})^b} = 2^{-b} e^{\kappa \psi} \int_0^\infty e^{-\omega \psi^2 / 2} p(\omega) \, d\omega,   
\end{align}
where \( \kappa = a - b/2 \) and \( \omega \sim \text{PG}(b, 0) \). A particular interest is when \( a = y \) and \( b = 1 \) and $\psi = \beta^\top x$, we recover an individual term of the logistic likelihood:
\[
p(y|\psi) = \frac{(e^{\psi})^y}{1 + e^{\psi}} = \frac{1}{2} e^{\kappa \psi} \int_0^\infty e^{-\omega \psi^2 / 2} p(\omega) \, d\omega,
\]
where \( \kappa = y - 1/2 \) and \( \omega \sim \text{PG}(1, 0) \).
Given a dataset of $N$ datapoints, we can now write down the likelihood as Gaussian 
conditioning on \( \omega \):
\begin{align}\label{eq:PG_Likeli}
p(\vy|\vpsi, \vomega) &\propto \prod_{i=1}^{N} e^{-\omega_i \psi_i^2 / 2} e^{\kappa_i \psi_i}, \nonumber\\ 
&\propto \mathcal{N}(\Omega^{-1} \vkappa | \psi, \Omega^{-1}),    
\end{align}
where \( \kappa_i = y_i - 1/2 \) and \( \Omega = \text{diag}(\vomega) \). 
Since we can write down the likelihood term as Gaussian when combined with the Gaussian prior over $\vpsi \sim \Nrm(\vmu, \Sigma)$, the posterior also becomes Gaussian conditioning on the PG variables $\vomega$:
\begin{align}\label{eq:posterior_f}
  p(\vpsi | \vy, \vomega) &\propto p(\vy|\vpsi, \vomega) p(\vpsi) \nonumber \\
  &\propto \mathcal{N}(\vpsi | \tilde{\Sigma} (\Sigma^{-1} \vmu + \vkappa) \tilde{\Sigma}) 
\end{align}
where \( \tilde{\Sigma} = (\Sigma^{-1} + \Omega)^{-1} \). 
The posterior inference is now analytically tractable given $\vomega$. To obtain the posterior distribution the PG variables $\vomega$, we can also employ the closed-form update given by \citep{PG13}:
\begin{align}\label{eq:posterior_omega}
    p(\vomega | \vy, \vpsi) &\propto \text{PG}(\vomega | 1, 0) e^{-\vomega \vpsi^2 / 2} \nonumber \\
    &\propto \text{PG}(\vomega | 1, \vpsi).
\end{align}
The Gibbs sampling procedure that alternates the two updates for $\vpsi$ in \autoref{eq:posterior_f} and $\vomega$ in \autoref{eq:posterior_omega} yields the joint posterior distributions over the function $\vpsi$ and the PG variables $\vomega$.

\vspace{-5pt}
\section{Method}
\vspace{-5pt}
Our method involves specifying the key components: (1) the likelihood, (2) the prior, and (3) the posterior distributions. We then explain how to perform posterior inference and parameter estimation. The complete algorithm is presented in \autoref{algo:pg_learning}. The subsequent sections provide a detailed explanation of each component.

\vspace{-2.5pt}
\subsection{Prior over the Logit Function}

\begin{algorithm}[!tb]
  \caption{One-vs-Each SoftMax Approximation with P\'olya-Gamma Auxiliary Variables for Prompt Learning}
  \label{algo:pg_learning}
  \begin{algorithmic}
    \STATE {\bfseries Input:} Objective
    $\mathcal{L}_{\text{elbo}} = \{\mathcal{L}_{\text{nll}}, \mathcal{L}_{\text{KLD}} \}$, Class prompts $\mathcal{C} \in \{c_1, \cdots, c_C \}$, The learnable prompts $\textbf{p}_{\vtheta_1}$, and residual prompts $\vr_{\vtheta_2}$,
    The predefined prompts for the original CLIP $\textbf{t} = \{ t_1, \cdots, t_C \}$, Image-Encoder $I$ and Text-encoder $T$,
    The number of parallel Gibbs chains $M$, Total number of classes $C$, Learning rate $\eta$, Weight for KLD $\beta$, $\alpha$ prior precision of $\vf$, Einstein summation convention, denoted by $\vpsi = \mA \vf$.
    \vspace{.1cm} 
    
    \STATE {\bfseries Initialize:} Parameters $\vtheta =\{\vtheta_1, \vtheta_2\}$ randomly.
    \REPEAT \STATE Sample a mini-batch $(\vx, \vy)$ from a training data
    $\Dat, \text{where} \ \vx \in \mathbb{R}^{n \times d}, \vy \in \{0,1\}^{n \times C}$
    
    \STATE $\mA \leftarrow {\text{OVE-MATRIX}}(\vy) \in \mathbb{R}^{C \times C \times C}$ in eq.(\ref{eq:A})

    $\vmu_{\vtheta} = I(\vx)^T T(\vp_{\vtheta_1}^c + \vr_{\vtheta_2}^c(\vx)) \in \mathbb{R}^{n \times C} $  
        
    $\vmu = I(\vx)^T T(\vp^c) \in \mathbb{R}^{n \times C} $

    \STATE $\vpsi_{\vtheta, nic} \leftarrow \sum_j \mA_{ijc} {\vmu_{\vtheta, ni}}, \quad \vpsi_{\vtheta} \in \mathbb{R}^{N \times C \times C}$
    \STATE $\vpsi_{nic} \leftarrow \sum_j \mA_{ijc} {\vmu}_{ni}, \quad \vpsi \in \mathbb{R}^{n \times C \times C} $  
    \STATE $\vkappa_{nic} = \sum_j \mA_{ijc} \big(\vy_{ni} - \nicefrac{1}{2}\big), \quad \vkappa \in \mathbb{R}^{n \times C \times C}$
    
    \FOR{$m=1$ {\bfseries to} $M$} 
    
        \STATE $\vomega^{(m)} \sim \text{PG}(1, \vpsi) \in \mathbb{R}^{n \times C \times C}$ \hfill given in eq.(\ref{eq:posterior_omega_PG})     

        \STATE $\vpsi^{(m)}_{\vtheta} \sim p(\vpsi^{(m)} | \vomega^{(m)}, \vpsi_{\vtheta}, \vkappa, \alpha )$ \hfill given in eq(.\ref{eq:posterior_f})

        \STATE $\vf^{(m)}_{\vtheta} \leftarrow \sum_C \vpsi^{(m)}_{\vtheta} \in \mathbb{R}^{n \times C}$ 
        \hfill due to $\vpsi = \bm{A} \vf_\vtheta  $

        \STATE $\mathcal{L}_{\text{nll}}^{(m)} \leftarrow - \sum_{i=1}^{N} \sum_{c=1}^{C} y_{i,c} \log f_{i,c}$  given in eq.(\ref{eq:OVE_likelihood})

        \STATE $\mathcal{L}_{\text{KLD}}^{(m)} \leftarrow \beta \lVert \vmu_{\vtheta} - \vmu  \rVert_2^2$

    \ENDFOR 

    \STATE $\theta \leftarrow \theta - \frac{\eta}{M} \sum_{m=1}^M \nabla_{\bf{\theta}} \big( \mathcal{L}_{\text{nll}}^{m} + \mathcal{L}_{\text{KLD}}^{m} \big)$
    
    \UNTIL{convergence}
  \end{algorithmic}
\end{algorithm}
\vspace{-2.5pt}
As mentioned earlier, we consider the logit function in the following form: 
\begin{align}
    f^{c}_\vtheta(\vx) = I(\vx)\trp T(\vp_\vtheta^{c} + \vr_\vtheta^{c}(\vx))
\end{align}
with frozen $I$ and $T$. In the following for the cleanness of the notation, we omit the $f$'s dependency on $\vtheta$, and simply write $f$ for $f_\vtheta$.

Given training data $\Dat=\{\mX, \mathbf{Y}\}$ which consists of input  points \( \mathbf{X} \in \mathbb{R}^{N \times D} \)
and corresponding one-hot labels \( \mathbf{Y} \in \{0, 1\}^{N \times C} \), where \( C \) is the number of classes, we denote
the logit function evaluated at those points as a single vector of length $N$
\begin{align}
    \mathbf{f}:= (f_1^1, \dots, f_N^1, f_1^2, \dots, f_N^2, \dots, f_1^C, \dots, f_N^C)^\top
\end{align}
and place an independent GP prior on the logits for each class: \( f_i^c(\mathbf{x}) \sim \mathcal{GP}(\mu(\mathbf{x}_i), k(\mathbf{x}_i, \mathbf{x}_j)) \).
The marginals on those points $\mX$ are jointly Gaussian 
    $p(\vf|\Dat) \propto \Nrm(\vf|\vmu, \mK)$
where \( \mu_i^c = \mu(\mathbf{x}_i) \) and \( \mathbf{K} \) is block diagonal with 
\( K_{ij}^c = k(\mathbf{x}_i, \mathbf{x}_j) \) for each block \( \mathbf{K}^c \).

To encode the information on the diverse sets of pre-training data into the model, we set $\vmu$ to the logit function of the pre-trained model, $\mu_i(\vx)=I(\vx_i)\trp T(\vp^c)$ and $\mK$ to an identity matrix with a small variance. This reflects that \textit{a priori} we believe that the best function value the model can take is the logit function from the pretrained model (because that model already saw so much of the world) with high confidence, where the prior precision is measured by a hyperparameter $\alpha$:
\begin{align}\label{eq:prior_f}
    \mbox{Prior: }  p(\vf|\Dat) & \propto \Nrm(\vf|\vmu, \alpha^{-1} \mI) 
\end{align} 

\subsection{Likelihood with the Logit Function}
The softmax likelihood under the multi-class classification setting with a function $f$ is given by 
\begin{align}\label{eq:softmax_likelihood}
   p(\vy_i|\vf(\vx_i)) {=} \frac{\exp[\vf^{\vy_i}(\vx_i)]}{\sum_{c}^C \exp[\vf^c(\vx_i)]}.
\end{align}
Here we make an \textit{one-vs-each (OVE)} approximation \citep{NIPS2016_814a9c18} to the likelihood is given by 
\begin{align}\label{eq:OVE_likelihood}
\mathcal{L}_{\text{OVE}}(\Dat|\vf) &=\prod_i \left( \prod_{c \neq c'} \sigma(\vf_i^{c} - \vf_i^{c'}) \right)^{\vy_i}
\end{align} 
See Supplementary material for detailed explanation on the OVE likelihood. 
Equipped with the OVE likelihood, we further add P\'olya-Gamma auxiliary variables $\vomega \in \mathbb{R}^N$ to the model.
Due to \autoref{eq:PG_Likeli}, we can represent the conditional likelihood of $\bm{\psi}$ (conditioned on $\bm{\omega}$) as Gaussian:
\begin{equation}
\resizebox{0.85\columnwidth}{!}{$
\begin{aligned}
  \mbox{Likelihood: }  \mathcal{L}_{OVE}(\bm{\psi} | \bm{\omega}, \Dat)
    &\propto \prod_{j=1}^{NC} e^{-\vomega_j \vpsi_j^2 / 2} \vkappa_j^{\psi_j} \\
    &\propto \mathcal{N}(\bm{\Omega}^{-1/2} \bm{\kappa} | \bm{\psi}, \bm{\Omega}^{-1})
\end{aligned}
$}
\label{eq:OVE_likelihood_2}
\end{equation}
where $\kappa_j = 1/2$, $\bm{\Omega} = \text{diag}(\bm{\omega})$ and $\bm{\psi}_{nic} = \sum_j \bm{A}_{ijc} \bm{f}_{ni}$ following the Einstein summation convention. For simplicity, we use the Einstein summation convention, denoted by $\vpsi = \mA \vf$ between tensors and matrices unless otherwise specified.   
Here, an OVE-matrix $\bm{A}$ is a $C \times C \times C$ sparse block matrix. $\bm{A}_c$ corresponds to a square matrix of dimension of $C$. Each block $\bm{A}_{c}$ matrix defined as follows:

\begin{equation}\label{eq:A}
    \bm{A}_{c} :=  \mathbf{1}_C \mathbf{e}_c^\top - \mathbf{I} 
\end{equation}
where, $\mathbf{1}_C$ is a column vector of dimension $C$ with all components equal to $1$ and the standard basis vector $\mathbf{e}_c = (0,\,0,\,\dots,\,0,\underbrace{1}_{c\text{-th}},\,0,\,\dots,\,0)^\top$. 
A simple example of computation $\mA$ and $\vf$ is provided in Appendix. 

%


\subsection{Posterior}

\begin{figure*}[!ht]
    \centering
    \includegraphics[width=0.92\linewidth]{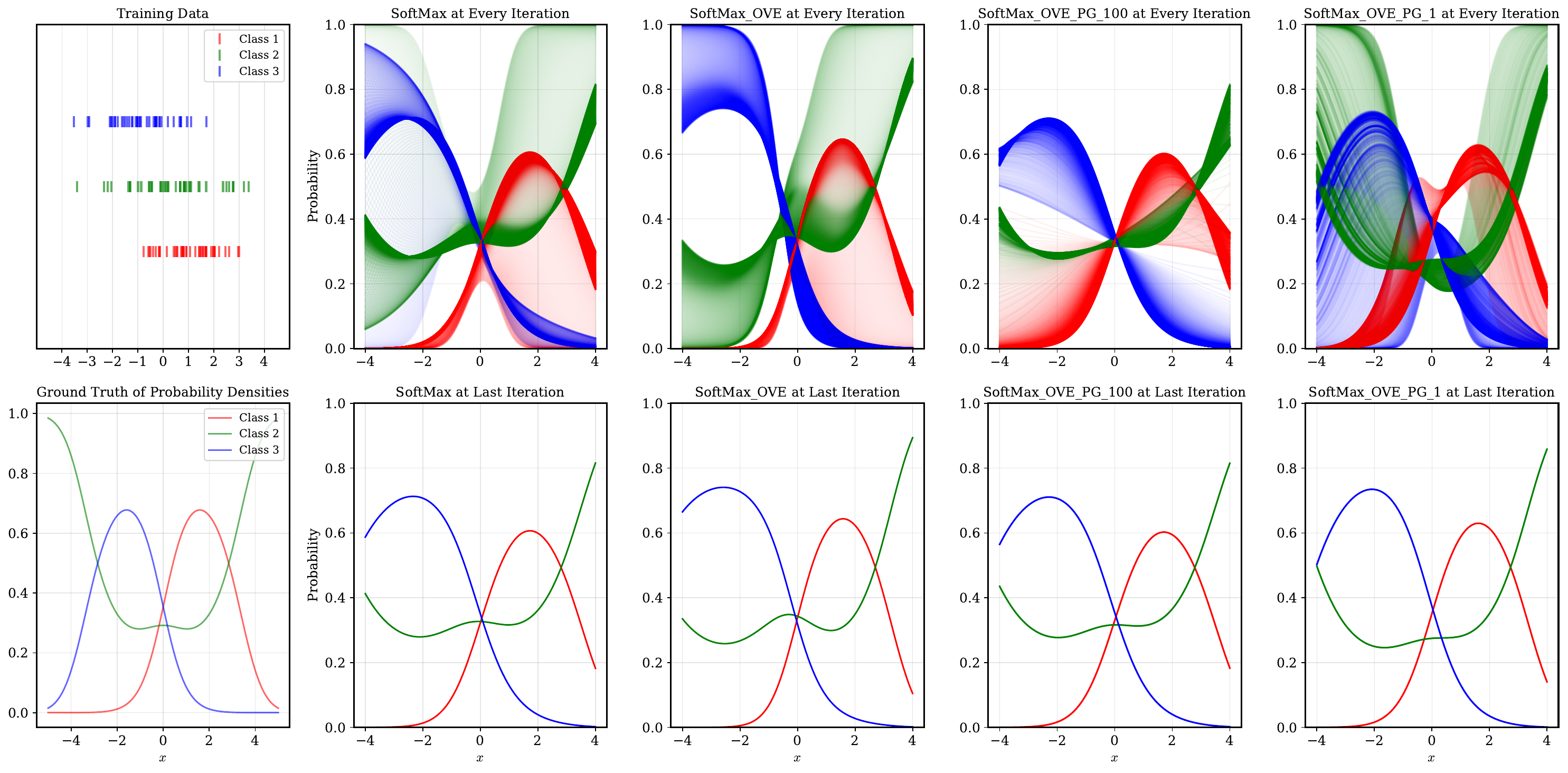}
    \vspace{-10pt}
    \caption{1D synthetic classification comparison between SoftMax, SoftMax\_OVE and SoftMAX\_OVE\_PG based on the gradient-descent. \textbf{[Column 1]} illustrates the arrangement of training samples \textbf{(top)} and the underlying ground-truth distributions \textbf{(bottom)}. \textbf{[Columns 2-5]} show the optimized functions from SoftMax, SoftMax\_OVE, and SoftMax\_OVE\_PG with the precision $\alpha$ values of 1 and 100, respectively. In these columns, the top row displays the class distributions at every iteration, while the bottom row shows the class  distribution at the last iteration.}
    \label{fig:1D_classification}
\end{figure*}

The exact posterior over $\vpsi$ with the OVE likelihood together with the P\'olya-Gamma data augmentation is given 
in \autoref{eq:posterior_f}.  This translates to the posterior over $\vf$ given by \citep{snell2021bayesian}
   \begin{align}\label{eq:correct_post_f}
   p(\vf|\vomega, \Dat) &\propto  \mathcal{L}_{\text{OVE}}(\vpsi|\vomega, \Dat) \Nrm(\vf|\vmu,\alpha^{-1} \mI), \nonumber \\
     &\propto \mathcal{N}(\bm{\Omega}^{-1/2} \bm{\kappa} | \bm{A} \vf, \bm{\Omega}^{-1}) \mathcal{N}(\vf | \bm{\mu},\alpha^{-1} \mI), \nonumber \\
     &\propto \mathcal{N}(\vf | \bm{\Sigma}(\alpha \bm{\mu} + \bm{A}^\top \bm{\kappa}), \bm{\Sigma})
\end{align} where $\bm{\Sigma} = (\alpha I + \bm{A}\trp \bm\Omega \bm{A})^{-1}$.
We modify this to an approximate posterior over $f$:
\begin{align}
    \label{eq:posterior_f_GP}
    \mbox{Posterior: } p(\vf|\vomega, \Dat) 
    & \approx \Nrm(\vf_\theta|\vmu_\theta,  \bm{\Lambda})
\end{align} 
where we set the mean function to the logit function of the prompt tuning model, i.e., $\vmu_\theta(\vx) = I(\vx)\trp T(\vp_\vtheta+\vr_\vtheta(\vx))$, and the covariance to $\Lambda = (\alpha \textbf{I} + \bm\Omega)^{-1}$. Notice that the mean function is the only quantity depending on the parameters of the prompt tuning model. Intuitively, sampling the posterior is guided by the dataset, adjusting logit values according to label information. Correct labels typically result in high logit values, reflecting higher confidence. Meanwhile, incorrect labels have lower logit values, indicating less confidence. We assume that the model learns to follow this behaviour during training. The PG distribution over the auxiliary variables is a conjugate prior for the given likelihood, so the posterior over the auxiliary variables is also PG distributed:
\begin{align}\label{eq:posterior_omega_PG}
        \mbox{Prior: }  p(\vomega) & \propto \prod_{i=1}^N\mbox{PG}(\vomega_i|1, 0) \nonumber \\
     \mbox{Posterior: } p(\vomega|\vf, \Dat) & \propto \mbox{PG}(\vomega|1, \mA \vf )
\end{align}
This posterior has to be updated whenever we have a new value for $\vf$. However, in our implementation, we simply set $\vf$ to the prior mean $\vmu$, where we set the prior mean $\vmu$ to the logit function of the pre-trained CLIP model.

\subsection{Parameter Estimation}
\label{subsec:param_est}
The ultimate goal is maximizing the likelihood of data after integrating out the unknowns $\vf$ and $\vomega$. We turn our attention to the variational lower bound using an approximating joint posterior over $\vf, \vomega$ denoted by $q(\vf, \vomega)$: 
\begin{align}\label{eq:variational_lower_bound}
&\log p(\Dat) \nonumber \\
&=\log \int_{\vomega} \int_{\vf}  p(\Dat| \vf, \vomega) p(\vf, \vomega) d \vomega d \vf, \nonumber \\
&= \log \int_{\vomega} \int_{\vf}  q(\vf, \vomega) p(\Dat| \vf, \vomega) \frac{p(\vf, \vomega)}{q(\vf, \vomega)}  d \vomega d \vf, 
\nonumber \\
&\geq \int_{\vomega} \int_{\vf} q(\vf, \vomega) \log \left( p(\Dat| \vf, \vomega) \frac{p(\vf, \vomega)}{q(\vf, \vomega)} \right)d \vomega d \vf, \nonumber \\
&\mbox{ Due to Jensen's inequality} \nonumber \\
& \quad = \int_{\vomega} \int_{\vf} q(\vf, \vomega) \log p(\Dat| \vf, \vomega)d \vomega d \vf \nonumber \\
& \qquad -D_{KL}[q(\vf, \vomega) || p(\vf, \vomega)].
\end{align}
Now we discuss each of these terms in \autoref{eq:variational_lower_bound}. 
The first term, expected log-likelihood of data, can be approximated via Monte Carlo integration using the samples of $\vf$ and $\vomega$:
\begin{align}\label{eq:first_term_loss}
&   \log \int_{\vomega} \int_{\vf}  p(\Dat|\vf, \vomega) q(\vf, \vomega) d \vomega d \vf \nonumber\\ 
&\approx \log \frac{1}{n}\sum_{i=1}^n \mathcal{L}_{\text{OVE}}({\vf_\vtheta}_i \mA \mid \vomega_i, \Dat),
\end{align} 
where the samples can be drawn from the posteriors given in \autoref{eq:posterior_f_GP} and \autoref{eq:posterior_omega_PG}. For speedy training, we can fix $\vomega$ to its posterior mean using \autoref{eq:posterior_omega_PG}, and sample only for $\mA \mathbf{f}^s_\vtheta$. Notice that $\vf^{s}_\vtheta$ is the only quantity that depends on the parameters of prompt learning, through $\vf^{s}_\vtheta \sim \Nrm(\vmu_\vtheta, \bm\Lambda)$.

\begin{figure*}[!ht]
    \centering
    \includegraphics[width=1.0\linewidth]{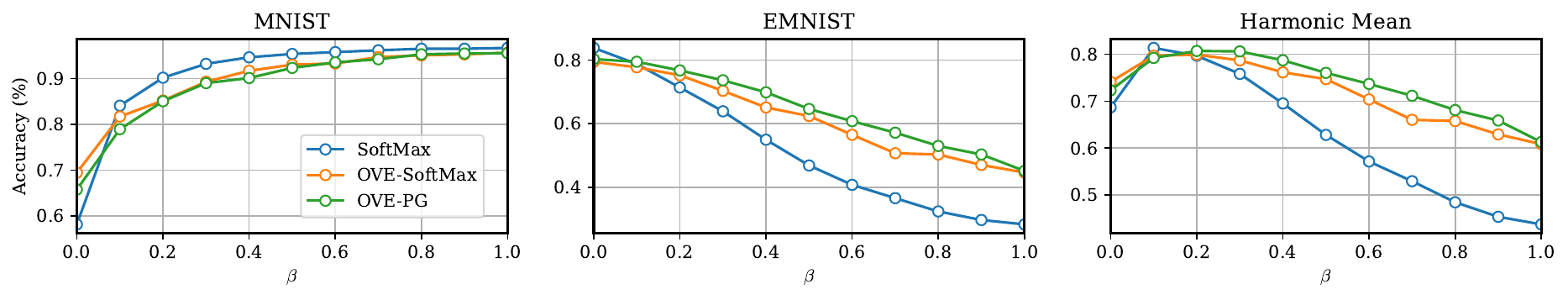}
    \caption{Effect of OVE-PG on ViT using the EMNIST and MNIST datasets. Our proposed OVE-PG 
    (\textcolor{green}{green line}) 
    achieves better generalization compared to Softmax 
    (\textcolor{blue}{blue line}) 
    and OVE-Softmax 
    (\textcolor{orange}{orange line}) 
    on unseen datasets (i.e., EMNIST), while slightly compromising performance on seen datasets (i.e., MNIST), resulting in higher overall performance.}
    \label{fig:vit}
\end{figure*}
For the second term, we first write down the joint distribution as  $q(\vf, \vomega) = q(\vf|\vomega)q(\vomega)$ and $p(\vf, \vomega)=p(\vf|\vomega) p(\vomega)$. In our prior formulation in \autoref{eq:prior_f}, the GP function $\vf$ is independent of the PG variable $\vomega$, yielding $p(\vf|\vomega) = p(\vf)$.
If we fix $\vomega$ to some value $\hat\vomega$, e.g., its posterior mean with \autoref{eq:posterior_omega_PG}, i.e., $\hat\omega = \mathbb{E}_{p(\vomega|\vf, \Dat)}[\vomega]$
and set $q(\vf|\hat\vomega)$ to the Gaussian posterior given in \autoref{eq:posterior_f_GP} and $p(\vf)$ to the Gaussian prior given in \autoref{eq:prior_f}, the second term becomes the KL divergence between posterior and prior distributions over $\vf$ which is in closed form:
\begin{align}\label{eq:KL}
   D_{KL}[q(\vf, \vomega) || p(\vf, \vomega)] 
    &\approx D_{KL}[q(\vf|\hat\vomega) || p(\vf)], \nonumber \\
    &\approx \|\vmu_\vtheta -\vmu \|_2^2,
\end{align} 
where in the last line, we further approximate the KL divergence between the two Gaussians by writing down only the term that depends on the parameters in prompt learning. The details of approximation can be found in \autoref{subsec_app_details}. 
To be able to control the strength of this regularization, we add a hyperparameter $\beta$. Our algorithm is summarized in \autoref{algo:pg_learning}. 

We construct the sampling procedure in the space of $\psi$ for computational efficiency. We assume no correlation between $\vf_{i}^c$ and $\vf_{i}^{c'}$. Given that $\mA$ is a sparse matrix and $\vpsi = \mA \vf$ is a deterministic projection, $\mA \vf $ represents the subtraction of two logit values, leading to doubling their variance. Additionally, when considering ${\mA}^T \vkappa$ in \autoref{eq:correct_post_f}, it reverts the one-hot label matrix and extend it along another axis by duplicating. For ${\mA}^T \bf{\Omega} {\mA}$, it influences the variance of the posterior distribution of $\vf$. Since ${\mA}$ is a sparse matrix with each row containing only two components, $1$ or $-1$, we approximate $\mA \approx \mI$. In this context, the components remain uncorrelated. Without additional modification, sampling in the $\vpsi$ space improves computational efficiency.
\paragraph{Empirical Validation with Synthetic Examples.} 
We compare OVE\_PG (\autoref{algo:pg_learning}) with SoftMax (\autoref{eq:softmax_likelihood}) and its OVE approximation (SoftMax\_OVE; \autoref{eq:OVE_likelihood}) in a synthetic 1D classification task. 
As shown in the second row of \autoref{fig:1D_classification}, all methods converge to the same result. However, their dynamic behaviour during training differs, as seen in the first row. Initially, SoftMax and its OVE approximation appear identical, but the P\'olya-Gamma variable and prior precision in OVE result in the extent of exploration. Specifically, when the precision $\alpha = 100$, the prior for each sigmoid function has a small variance, while $\alpha = 1$ results in a larger variance, promoting greater exploration during training. This exploration helps avoid overfitting and improves performance. The detailed procedure is provided in Appendix.

\paragraph{Empirical Validation with ViT.}
We also evaluate our proposed method on image classification using ViT \citep{dosovitskiy2021an}. The ViT model is first trained on MNIST with SoftMax and then fine-tuned on EMNIST with SoftMax, OVE, and OVE-PG, respectively. We tested on MNIST, EMNIST and the harmonic mean of both datasets. A sensitivity analysis on the hyperparameter $\beta$ illustrates how performance changes with different values. Details of the experimental setup are in the Appendix. As shown in \autoref{fig:vit}, OVE-PG consistently outperforms across most $\beta$ values, while softmax shows a notable performance drop on EMNIST and the harmonic mean as $\beta$ increases.

\vspace{-5pt}
\section{Related Work}
In this section, we review relevant literature related to our work, including vision-language models, prompt learning for such models, and Bayesian inference for multi-class.

\paragraph{Vision-Language Models (VLMs).} VLMs address the shortcomings of vision-only supervised learning by introducing robustness and flexibility in zero-shot inference through natural language supervision. A pioneering work in this field is CLIP \citep{radford2021learning}, which employs contrastive learning on a massive dataset of 400 million image-text pairs. Building on this foundation, ALIGN \citep{jia2021scaling} further scales up the dataset with an even larger collection of noisy image-text pairs to enhance performance. CoCa \citep{yu2022coca} integrates both captioning and contrastive losses, effectively combining the strengths of contrastive methods like CLIP with generative approaches. EVA-CLIP \citep{sun2023eva} applies various training strategies, including alternative attention mechanisms and optimizers, to further boost CLIP's performance. Additionally, SigLIP \citep{zhai2023sigmoid} replaces the traditional softmax loss with a sigmoid loss, enabling more efficient pretraining with smaller batch sizes.


\begin{table*}[t]
    \centering
    \resizebox{0.97\linewidth}{!}{
    \begin{tabular}{l|cc|cc|cc|cc}
    \toprule[0.1em]
                        & \multicolumn{2}{c|}{CoOp} & \multicolumn{2}{c|}{CoCoOp} & \multicolumn{2}{c|}{MaPLe} & \multicolumn{2}{c}{APEX} \\
                        &   SoftMax    &  OVE-PG   &   Softmax   &   OVE-PG  &   SoftMax    &  OVE-PG &   SoftMax    &  OVE-PG \\
        \midrule
        Caltech101      &   94.43   &   \textbf{94.93}   &   94.03   &   \textbf{94.13} & 94.00 & \textbf{94.55} & \textbf{95.00} & 94.77  \\
        DTD             &   54.97   &   \textbf{55.67}   &   55.07   &   \textbf{56.17} & 50.97 & \textbf{57.33} & 58.67 & \textbf{61.13}   \\
        EuroSAT         &   55.20   &   \textbf{71.23}   &   66.63   &   \textbf{68.03} & 60.70 & \textbf{63.43} & 68.23 & \textbf{71.37} \\
        FGVC Aircraft   &   22.93   &   \textbf{36.23}   &   \textbf{34.13}   &   33.60 & 34.17 & \textbf{36.97} & 34.83 & \textbf{35.13}   \\
        Food101         &   91.70   &   \textbf{91.80}   &   \textbf{91.73}   &  91.70  & 91.40 & \textbf{91.60} & \textbf{91.80} & 91.73  \\
        Flowers102      &   72.47   &   \textbf{73.83}   &   74.07   &   \textbf{74.33} & 73.13 & \textbf{75.60} & 73.37 & \textbf{74.67}   \\
        Oxford Pet      &   \textbf{97.77}   &  97.70    &  \textbf{97.63}   &   96.47  & \textbf{97.87} & 97.63 & \textbf{96.83} & 96.80 \\
        Stanford Cars   &   72.73   &   \textbf{75.23}   &   74.70   &   \textbf{75.50} & 72.40 & \textbf{74.07} & \textbf{74.50} & 74.40   \\
        SUN397          &   75.90  &    \textbf{77.53}   &   76.87   & \textbf{77.40}  & 76.77 & \textbf{78.40} & \textbf{76.97} & 76.87  \\
        UCF101          &   72.43   &   \textbf{76.20}   &   72.63   &   \textbf{73.27} & 76.40 & \textbf{78.47} & 74.77 & \textbf{75.23}    \\
        \midrule
        \textit{Average (Unseen)}&   71.05   &   \textbf{75.04}   &   73.75   &   \textbf{74.06}  & 72.78 & \textbf{74.91} & 74.50 & \textbf{75.21} \\
    \bottomrule[0.1em]
    \end{tabular}}
    \caption{Comparison of accuracy (\%) on unseen classes between Softmax and OVE-PG (Ours) for CoOP, CoCoOp, MaPLe, and APEX. The best performance within each shared base algorithm is highlighted in \textbf{bold}.}
    \label{tab:unseen}
\end{table*}
\paragraph{Prompt Learning in VLMs.} Parameter-efficient fine-tuning with soft prompts, originally from natural language processing, has gained attention \citep{lester2021power}. This technique has been applied in vision-language models to adapt to downstream tasks. CoOp \cite{zhou2022learning} first introduced learnable prompts for CLIP, replacing manual ones. CoCoOp conditions text prompts on images to prevent overfitting \cite{zhou2022conditional}, while KgCoOp minimizes the gap between learned and manual prompts \citep{yao2023visual}. ProGrad aligns gradient directions to retain general knowledge \citep{zhu2022prompt}, and PromptSRC uses multiple regularization losses with Gaussian aggregation of weights to prevent forgetting \citep{khattak2023self}. Recent studies reported that visual learnable prompts significantly enhance generalization capability \citep{khattakMaPLe, yang-etal-2024-towards}.
\citet{Derakhshani_2023_ICCV} proposed a Bayesian approach to reduce overfitting by introducing an additional network. In contrast, our method shows that a simple Bayesian approach can effectively mitigate overfitting without requiring extra regularization or network parameters.


\vspace{-2.5pt}
\paragraph{Bayesian Inference for Multi-Class.}
The logistic function is robust in handling outliers for classification tasks, but maintaining conjugacy between prior and posterior distributions is challenging. Initially, for binary classification, the sigmoid function was modeled as a Gaussian distribution conditioned on a variable from the P\'olya-Gamma distribution \citep{PG13}. However, this approach doesn’t naturally extend to the softmax function. Inspired by Dirichlet processes, the stick-breaking concept has been applied to transform multi-class classification into a series of binary classifications, distinguishing each class from the others \citep{linderman2015dependent}. Since this method builds on sequence dependency, changing the order leads a different outcome. To address this, the softmax function is decomposed into several sigmoid functions based on the composite likelihood principle \citep{NIPS2016_814a9c18}, each retaining conjugacy with P\'olya-Gamma variables \citep{snell2021bayesian}. While this technique showed theoretical and empirical results in few-shot classification tasks, its reliance on Gaussian processes is a limitation. Specifically, it introduces significant computational burdens as the number of classes and data points increases. Due to this complexity, this approach typically assumes a maximum of five classes and data points to maintain feasibility. In contrast, the proposed method builds on this framework but leverages prior knowledge from a pre-trained VLM. By sampling an appropriate logit function with an approximated softmax, it reduces computational complexity while retaining the original behaviour.

\section{Experiments}
Here, we present the experimental results that verify the effectiveness of our proposed method. We begin by introducing the experimental setup, followed by the relevant results for image recognition using CLIP.

\subsection{Experimental Setup}
We evaluate our proposed method on the unseen prompts generalization task, focusing on 10 image recognition datasets, following \citet{zhou2022conditional}. These datasets cover multiple recognition tasks, including Caltech101 \citep{FeiFei2004LearningGV} which consists of generic objects; Oxford Pets \citep{parkhi2012cats}, FGVC Aircraft \citep{maji2013fine}, Flowers102 \citep{nilsback2008automated}, Food101 \citep{Bossard2014Food101M}, and Stanford Cars \citep{krause20133d} for fine-grained classification, UCF101 \citep{soomro2012dataset} for action recognition, SUN397 \citep{Xiao2010SUNDL} for scene recognition, UCF101 \citep{Soomro2012UCF101AD} for action recognition, DTD \citep{cimpoi2014describing} for texture classification, and EuroSAT \citep{helber2019eurosat}, which consists of satellite images.

We integrate our proposed method into CoOp \citep{zhou2022learning}, CoCoOp \citep{zhou2022conditional}, MaPLe \citep{khattakMaPLe} and APEX \citep{yang-etal-2024-towards}, as well as compare them to the baseline versions with the vanilla softmax function. In this experiment, we pay attention to how effectively the learned prompts generalize to unseen classes or domains. We utilize the ViT-B/16 model as the CLIP image encoder and a standard GPT2-like structure with an End-Of-Text (EOT) token as the classification token for the text encoder. For training, we follow the approach outlined in \citet{Derakhshani_2023_ICCV}, except for using a batch size of 4. We use the SGD optimizer with a learning rate of 0.002 and a cosine learning rate scheduler. Also, all evaluations are conducted for three different seeds. The detailed experimental setup can be found in Appendix.

\subsection{Main Results}
\paragraph{Seen-to-Unseen Generalization.} We evaluate the generalization performance of our proposed method, OVE-PG, on 10 datasets, each tested with multiple random seeds. In each dataset, classes are split into seen and unseen subsets. Our model is trained on the seen classes and evaluated on the unseen classes. As shown in the \autoref{tab:unseen}, OVE-PG consistently outperforms the softmax baseline for all existing approaches. Notably, OVE-PG with CoOp achieves the highest gap (+3.99\%) among all cases, which indicates that our regularization effect sufficiently alleviates overfitting on seen classes. Also, OVE-PG provide consistent performance improvement across all baselines, including CoCoOp (+0.31\%), MaPLe (+2.13\%), and APEX (+0.71\%), reinforcing its effectiveness in generalizing to unseen classes. Additionally, we report the performance on seen categories and the harmonic mean between seen and unseen categories provided in Appendix. Consequently, while OVE-PG introduces slight performance degradation in seen classes, it achieves higher or comparable performance from the perspective of the harmonic mean, which considers both seen-and-unseen generalization.

\paragraph{Cross-dataset Evaluation.} To enable the model to generalize across different domains, we employed a cross-dataset evaluation task. Specifically, we first trained the model on ImageNet \citep{deng2009imagenet} and then transferred it to six other datasets. As shown in \autoref{table:cross_eval}, OVE-PG achieves higher average performances compared to Softmax baselines. Notably, OVE-PG outperforms in four out of six datasets for both CoOp and CoCoOp cases, while only slightly compromising or even improving performance on the source dataset. This demonstrates the effectiveness of our proposed method, especially in challenging situations where both the task and domain are unseen.

\subsection{Analysis}
\paragraph{Component Analysis.} To demonstrate the effectiveness of our PG augmentation, we compare OVE-PG to OVE without PG augmentation, as reported in \autoref{tab:component}. This indicates our performance improvements are primarily due to the novel PG augmentation, rather than the OVE method itself. Notably, while OVE performed worse on datasets where CoOp already struggled (e.g., DTD, EuroSAT, and FGVC Aircraft), OVE-PG significantly outperforms the baselines. Additionally, OVE-PG surpasses OVE even in datasets where OVE performs well, such as UCF101.

\begin{table}[t]
\centering
\resizebox{1.0\columnwidth}{!}{
\begin{tabular}{l|l|cc|cc}
\toprule[0.15em]
 & & \multicolumn{2}{c|}{CoOp} & \multicolumn{2}{c}{CoCoOp} \\
 & & SoftMax & OVE-PG & SoftMax & OVE-PG \\
\midrule
\textbf{Source} & ImageNet & \textbf{76.5} & 74.5 & 75.9 & \textbf{76.1}  \\ \midrule
\multirow{6}{*}{\textbf{Target}} 
 & DTD & 51.7 & \textbf{57.4} & 54.5 & \textbf{55.4} \\
 & EuroSAT & 51.3 & \textbf{61.7} & 52.5 & \textbf{55.6} \\
 & Aircraft & \textbf{26.7} & 23.8 & 24.2 & \textbf{26.8} \\
 & Flower102 & \textbf{72.6} & 71.5 & \textbf{73.8} & 71.5  \\
 & Cars & 62.8 & \textbf{63.6} & \textbf{65.2} & 64.9  \\
 & UCF101 & 67.2 & \textbf{69.9} & 69.2 & \textbf{70.6} \\
 \midrule
 \multicolumn{2}{c|}{\textit{Average}} & 58.40 & \textbf{60.34} & 59.56 & \textbf{60.13} \\
\bottomrule[0.12em]
\end{tabular}
}
\caption{Comparison of accuracy (\%) on cross-dataset tasks between Softmax and OVE-PG (Ours) for CoOP and CoCoOp. The best performance within each shared base algorithm is highlighted in \textbf{bold}.}
\label{table:cross_eval}
\end{table}
\begin{table}[t]
    \centering
    \resizebox{1.0\columnwidth}{!}{
    \begin{tabular}{l|cccc}
    \toprule[0.15em]
                        &   DTD    &  EuroSAT   &   FGVC Aircraft   &   UCF101 \\
        \midrule
        SoftMax         & 54.97 & 55.20 & 22.93 & 72.43 \\ \midrule
        \multirow{2}{*}{OVE}   & 50.93 & 54.93 & 16.97 & 76.03 \\
                               & (\textcolor{blue}{-4.04\%}) & (\textcolor{blue}{-0.27\%}) & (\textcolor{blue}{-5.96\%}) & (\textcolor{red}{+3.60\%}) \\ \midrule 
        \multirow{2}{*}{OVE-PG}& \textbf{55.67} & \textbf{71.23} & \textbf{36.23} & \textbf{76.20} \\
        & (\textcolor{red}{+0.70\%}) & (\textcolor{red}{+16.03\%}) & (\textcolor{red}{+13.30\%}) & (\textcolor{red}{+3.77\%}) \\
    \bottomrule[0.12em]
    \end{tabular}}
    \caption{Comparison of accuracy (\%) on unseen classes between OVE and OVE-PG (Ours) for CoOP. The results indicate that our PG augmentation significantly improve the performance of VLMs.}
    \label{tab:component}
\end{table}
\begin{figure*}[t]
    \centering
    \includegraphics[width=1.0\linewidth]{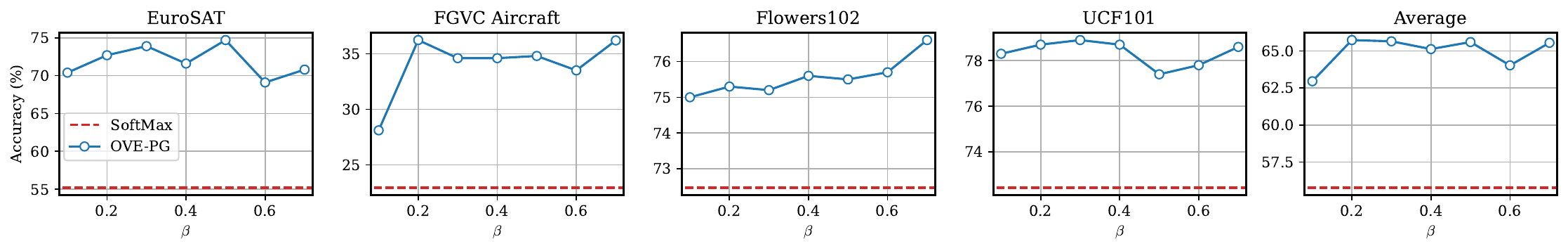}
    \vspace{-10pt}
    \caption{Sensitivity analysis of the hyperparameter $\beta$ for controlling KLD regularization in OVE-PG (\textcolor{blue}{blue solid line}), compared to the softmax baseline for CoOp (\textcolor{red}{red dotted line}). The results demonstrate robustness within a moderate range of $\beta$ values ($0.2$–$0.5$) across all datasets.}
    \label{fig:sensitivity}
    \vspace{-5pt}
\end{figure*}
\paragraph{Sensitivity Analysis.} The main hyperparameter for our proposed method is $\beta$, which controls the strength of the KL divergence between the pre-trained and fine-tuned distributions. To understand the effect of this hyperparameter, we conducted a sensitivity analysis on $\beta$ using CoOp. \autoref{fig:sensitivity} summarizes the results for controlling different levels of KD regularization on four datasets: EuroSAT, FGVC Aircraft, Flowers102, UCF101, as well as the average results across these datasets. Our OVE-PG consistently outperforms the baseline by large margins across all values of $\beta$, ranging from $0.1$ to $0.7$. Notably, within the mild range of $\beta$ (i.e., $0.2$–$0.5$), OVE-PG shows robustness regardless of the specific choice of $\beta$, which may alleviate the effort required for tuning this hyperparameter. Extended results can be found in Appendix.

\vspace{-5pt}
\section{Conclusion and Discussion}
\vspace{-5pt}
In this paper, we have proposed a Bayesian framework, coined as OVE-PG, that integrates P\'olya-Gamma augmentation with a one-vs-each softmax approximation to mitigate overfitting in prompt learning for VLMs. OVE-PG outperforms existing methods by balancing task-specific learning with global knowledge retention. Notably, it does so without requiring additional regularization terms or network parameters, maintaining simplicity and compatibility with existing models. Our theoretical and experimental results demonstrate the robustness of this approach across various datasets. Additionally, we verify that our method is effective not only for VLMs but also in simple 1D examples and image classification with ViT. We believe that our approach provides a promising direction for improving prompt learning in VLMs, and hope that our framework can be expanded in a wide range of other domains.


\subsubsection*{Acknowledgements}
M. Kim and M. Park were supported in part by the Natural Sciences and Engineering Research Council of Canada (NSERC) and the Canada CIFAR AI Chairs program at AMII (Alberta Machine Intelligence Institute). We also thank our anonymous reviewers for their constructive feedback, which has helped improve our manuscript.

\bibliography{main}
\bibliographystyle{apalike}

\clearpage
\appendix
\setcounter{equation}{0}
\renewcommand{\theequation}{\Alph{section}.\arabic{equation}}
\setcounter{table}{0}
\renewcommand{\thetable}{\Alph{section}.\arabic{table}}
\setcounter{figure}{0}
\renewcommand{\thefigure}{\Alph{section}.\arabic{figure}}

\onecolumn
\aistatstitle{
Supplementary Material:\\
Bayesian Principles Improve Prompt Learning 
In VLMs}

\section{One-vs-Each SoftMax Approximation}
We provide a detailed proof of one-vs-each softmax approximation \citep{NIPS2016_814a9c18}. This serves as a lower bound to the softmax likelihood. In this section, we assume there are $n$ data points and define $\vf \in \mathbb{R}^{n \times \mathcal{C}}$ and $\vy = \mathbb{R}^{n \times \mathcal{C}}$. For simplicity, we focus on a single data point, denoted as $\vy_i = \{y_i^1, \cdots, y_i^\mathcal{C} \}$ and $\vf_i = \{f_i^1, \cdots, f_i^\mathcal{C} \}$. Starting with the softmax likelihood, we have,
\begin{equation}
    \begin{aligned}
    \label{eq_softmax}
        p(\vy_i = c \mid \vf_i) = \frac{e^{f_i^c}}{\sum_{c^{'}} e^{f_i^{c'}}} = \frac{1}{1 + \sum_{c' \neq c} e^{-(f_i^{c} - f_i^{c'})}}
    \end{aligned}
\end{equation}

To derive the OVE approximation, we manipulate the denominator by separating the term  $e^{f_i}$ from the sum as follows.

\begin{equation}
    \begin{aligned}
    \label{eq_denominator_softmax}        
\sum_{c=1}^\mathcal{C} e^{f_i^c} = e^{f_i^c} + \sum_{c' \neq c} e^{f_i^{c'}}
    \end{aligned}
\end{equation}

Dividing both the numerator and the denominator by  $e^{f_i}$, the softmax probability becomes,

\begin{equation}
    \begin{aligned}
    \label{eq_softmax_rewrite}        
p(\vy_i = c \mid \vf_i) &= \frac{e^{f_i^c}}{e^{f_i^c} + \sum_{c' \neq c} e^{f_i^{c'}}} \\
&= \frac{1}{1 + \sum_{c' \neq c} e^{f_i^{c'} - f_i^{c}}} \\
&= \frac{1}{1 + \sum_{c' \neq c} e^{-(f_i^{c} - f_i^{c'})}}
    \end{aligned}
\end{equation}

This expression highlights the dependence on the differences  $f_i^c - f_i^{c'}$. To establish a lower bound, we utilize the following general inequality, which holds for any set of non-negative real numbers  $\{ \alpha_k \}$,

\begin{equation}
    \begin{aligned}
    \label{eq_general_inequality}        
1 + \sum_{k} \alpha_k \leq \prod_{k} (1 + \alpha_k) \quad \text{such that} \quad \alpha_k \geq 0
    \end{aligned}
\end{equation}

Applying this inequality to the denominator in \autoref{eq_softmax_rewrite}, with  $\alpha_{c'} = e^{-(f_i^c - f_i^{c'})} \geq 0$, we obtain

\begin{equation}
    \begin{aligned}
    \label{eq_recipocal_ove_softmax_inequality}        
1 + \sum_{c' \neq c} e^{-(f_i^{c} - f_i^{c'})} \leq \prod_{c' \neq c} \left(1 + e^{-(f_i^c - f_i^{c'})}\right)
    \end{aligned}
\end{equation}

Since all terms are positive, by taking reciprocals on both sides, we have

\begin{equation}
\begin{aligned}
\label{eq_reciprocal_inequality}
\frac{1}{1 + \sum_{c' \neq c} e^{-(f_i^{c} - f_i^{c'})}} \geq \frac{1}{\prod_{c' \neq c} \left(1 + e^{-(f_i^{c} - f_i^{c'})}\right)}
\end{aligned}
\end{equation}

Recognizing that $\nicefrac{1}{1 + e^{-x}}$ is the sigmoid function  $\sigma(x)$, we can express each term in the product as:

\begin{equation}
\begin{aligned}
\label{eq_sigmoid_relation}
\frac{1}{1 + e^{-(f_i^{c} - f_i^{c'})}} = \sigma(f_i^c - f_i^{c'})
\end{aligned}
\end{equation}

Therefore, the product in the denominator becomes,

\begin{equation}
\begin{aligned}
\label{eq_product_sigmoids}
\prod_{c' \neq c} \frac{1}{1 + e^{-(f_i^c - f_i^{c'})}} = \prod_{c' \neq c} \sigma(f_i^c - f_i^{c'})
\end{aligned}
\end{equation}

Combining \autoref{eq_reciprocal_inequality} and \autoref{eq_product_sigmoids}, we establish the following inequality,

\begin{equation}
\begin{aligned}
\label{eq_ove_lower_bound}
p(\vy_i = c \mid \vf_i) \geq \prod_{c' \neq c} \sigma(f_i^c - f_i^{c'})
\end{aligned}
\end{equation}

\section{Polya-Gamma Augmentation for Sigmoid Function}
\label{app_polya_gamma}

\citet{PG13} introduced the Polya-Gamma augmentation for the sigmoid function to facilitate Bayesian inference in logistic linear regression for binary classification. While previous methods approximate the likelihood function using Gaussian distributions, this approach enables a natural derivation of the likelihood as a Gaussian distribution by introducing augmented variables drawn from the Polya-Gamma distribution. The detailed procedure is provided as follows.

\subsection{Sigmoid Function}
We denote input $\vx_i \in \mathbb{R}^D$ and $y_i \in \{0, 1\}$. The number of data points is $N$.
The sigmoid function is denoted by $\sigma(f_i) = \frac{1}{1+\exp(-f_i)}$ and $f_i = \vbeta^T\vx_i$. We assume a linear model parameterized by $\vbeta$.

\begin{equation}
    \begin{aligned}
    \label{eq_sigmoid_function}
    p(\textbf{y}|f) &= \prod_{i =1}^{N} \sigma(f_i)^{y_i} (1 - \sigma(f_i))^{1 - y_i} \\
                    &= \prod_{i =1}^{N} \left( \frac{1}{1+e^{-f_i}} \right)^{y_i} \left( \frac{e^{-f_i}}{1+e^{-f_i}} \right)^{1 - y_i} \\
                    &= \prod_{i =1}^{N} \frac{1}{(1+e^{-f_i})^{y_i}} \frac{{e^{-f_i}}^{(1-y_i)}}{(1+e^{-f_i})^{1 - y_i}} \\
                    &= \prod_{i =1}^{N} \frac{{e^{-f_i}}^{(1-y_i)}}{1+e^{-f_i}} = \prod_{i =1}^{N} \frac{({e^{-(\vbeta^T\vx_i)}})^{(1-y_i)}}{1+e^{-\vbeta^T\vx_i}}
    \end{aligned}
\end{equation}

\subsection{Understanding $\cosh(x)$}
To demonstrate the relationship between $\cosh(x)$ and $\prod_{k=1}^{\infty} \big( 1 + \frac{t}{2\pi^2(k-\frac{1}{2})^2} \big)$,
we first explore the structure of infinite product representation of $\sin(\pi x)$ and $\cos(\pi x)$. 
A key step in this process is to examine the zeros of hypoerbolic cosine function $\cosh(x) = 0$. 
We begin by considering the following equality:

\begin{equation}
    \begin{aligned}
    \label{eq_zero_cosh}
    \cosh(x) = 0 &\rightarrow \frac{e^x + e^{-x}}{2} = 0 \rightarrow e^{2x} + 1 = 0 &\rightarrow e^{-2x} = -1 \\
    \end{aligned}
\end{equation}
The solution of \autoref{eq_zero_cosh} lies in the imagenary domains because of $e^{-2x} = -1$.
To explore this, we substitute $x = iy$, where $y$ is real-value. Under this substitution, $\cosh(iy) = \frac{e^{iy} + e^{-iy}}{2} = \cos(y)$, 
therefore $\cosh(x) = 0$ when $\cos(y) = 0$. 
The zeros of cosin function occur at $y = (n + \frac{1}{2}) \pi$ for $n \in \mathbb{Z}$ because $\cos(\frac{1}{2}\pi)=0, \cos(\frac{3}{2}\pi)=0, \cdots$.
So, we rewrite the zeros of $\cosh(x)$ as $x = i \pi (n + \frac{1}{2})$ for $n \in \mathbb{Z}$.

Next, we consider the infinite product representation of $\cosh(x)$. 
According to the definition of entrie function and Weierstrass factorization theorem, any entire function $f(z)$ can be represented as an infinite product of its zeros $z_k$.
In details, it is described as follows. 
\begin{equation}
    \begin{aligned}
    \label{eq_Weierstrass_factorization}
    f(z) = e^{g(z)} \prod_{n=1}^{\infty} \left(1 - \frac{z}{z_n}\right) e^{\frac{z}{z_n} + \frac{1}{2}\left(\frac{z}{z_n}\right)^2 + \cdots + \frac{1}{m_n}\left(\frac{z}{z_n}\right)^{m_n}} \\
    \end{aligned}
\end{equation}

By entire function of order 1 and symetric property of $\cosh(x)$ about the origin (zero), the product simplifies and the exponential factors cancel out.
We can write $ \cosh(x) = C \prod_{n=1}^{\infty} \left(1 - \frac{x^2}{x_n^2}\right) $ where, $C$ is constant. 
To simplify the infinite product, we substitute $x_n = i \pi (n + \frac{1}{2})$ 
into $x_n^2 = -\left(n + \tfrac{1}{2}\right)^2 \pi^2$.
As a result, we have $1 - \frac{x^2}{x_n^2} = 1 - \frac{x^2}{-\left(n + \tfrac{1}{2}\right)^2 \pi^2} = 1 + \frac{x^2}{\left(n + \tfrac{1}{2}\right)^2 \pi^2}$.
Therefore, the product becomes $\cosh(x) = C \prod_{n=0}^{\infty} \left(1 + \frac{x^2}{\pi^2 \left(n + \tfrac{1}{2}\right)^2}\right)$.
In this context, assuming $C = 1$ simplifies the expression further and $n=k-1 \ \ \text{s.t} \ \ k>1$ :
\begin{equation}
    \begin{aligned}
    \label{eq_coshx_infinite_product}
    \cosh(x) = \prod_{k=1}^{\infty} \left( 1 + \frac{x^2}{\pi^2 (k - \frac{1}{2})}  \right) \\
    \end{aligned}
\end{equation}

\subsection{Laplace Transformation and Moment Generating Function}
The Laplace transformation $\mathcal{L}$ is defined as follows:

\begin{equation}
    \begin{aligned}
        \label{eq_laplace_transform}
        \mathcal{L}\{f(\omega) \}(t) &= \int_{0}^{\infty} e^{-\omega t} f(\omega) d\omega \\
    \end{aligned}
\end{equation}
Suppose that the moment generating function (MGF) of $x$ is given by $\mathcal{M}_{x}(t) = \mathbb{E} \big[e^{tx}  \big]$.
For instance, the MGF of Bernoulli distribution is given by $M_{x}(t) = 1 - p + pe^t$.
Assuming that $f(\omega)$ is a probability density function, the Laplace transformation of $f(\omega)$ is the MGF of $x$.
\begin{equation}
    \begin{aligned}
        \label{eq_mgf}
            \mathcal{L}\{f(\omega) \}(t) &= M_{\omega}(-t) = \mathbb{E} \big[e^{-t\omega}  \big]
    \end{aligned}
\end{equation}
For instance, the moment generating function of gamma distribution with the shape parameter $b$ and scale parameter $1$ is known as $\mathcal{M}_{\gamma}(t) = \mathbb{E}_{\gamma}\left[ e^{t \gamma} \right] = (1 - t)^{-b}$ 
and its laplace transformation is given by $\mathcal{L}_{p(\gamma)}(t) = (1 + t)^{-b}$.

Here is the derivation of the MGF of gamma distribution with shape parameter $b$ and scale parameter $1$.
By definition of MGF and gamma distribution, we have:

\begin{equation}
    \begin{aligned}
        \label{eq_mgf_gamma}
            M_x(t) = \int_{0}^{\infty} e^{t x} \frac{x^{b - 1} e^{-x}}{\Gamma(b)} \, dx = \frac{1}{\Gamma(b)} \int_{0}^{\infty} x^{b - 1} e^{-(1 - t)x} \, dx
    \end{aligned}
\end{equation}
Note that the interal in \autoref{eq_mgf_gamma} converges only if $(1 - t) > 0$, i.e., $t < 1$.
Next, recall the definition of the gamma function $\Gamma(\alpha) = \int_{0}^{\infty} x^{\alpha - 1} e^{-x} dx$,
our integral in \autoref{eq_mgf_gamma} has a similar form but with $e^{-(1 - t)x}$  instead of  $e^{-x}$.
To align it with the Gamma function, we introduce the substitution $\lambda = 1 - t$ for \autoref{eq_mgf_gamma}, where $\lambda > 0$  when  $t < 1$. 
Then, we have $M_x(t) = \frac{1}{\Gamma(b)} \int_{0}^{\infty} x^{b - 1} e^{-\lambda x} dx$.
This integral is the definition of the Gamma function scaled by $\lambda$:
\begin{equation}
    \begin{aligned}
        \label{eq_gamma}
        \int_{0}^{\infty} x^{b - 1} e^{-\lambda x} \, dx = \frac{\Gamma(b)}{\lambda^b}
    \end{aligned}
\end{equation}

Finally, we organize the results and substitute back to the MGF. Consider $\int_{0}^{\infty} x^{b - 1} e^{-\lambda x} \, dx = \lambda^{-b} \int_{0}^{\infty} (\lambda x)^{b - 1} e^{-\lambda x} \, d(\lambda x) \cdot \frac{1}{\lambda}$,
To write this in a form of gamma distribution, we have substitution $d(\lambda x) = \lambda dx$ then we obtain:
\begin{equation}
    \begin{aligned}
        \label{eq_gamma_final_1}
            \lambda^{-b} \int_{0}^{\infty} y^{b - 1} e^{-y}dy = \lambda^{-b} \Gamma(b)
    \end{aligned}
\end{equation}
Therefore,
\begin{equation}
    \begin{aligned}
        \label{eq_gamma_final_2}
            \int_{0}^{\infty} x^{b - 1} e^{-\lambda x}dx = \frac{\Gamma(b)}{\lambda^b}
    \end{aligned}
\end{equation}
Now substitute back:
\begin{equation}
    \begin{aligned}
        \label{eq_mgf_gamma_final}
            M_x(t) &= \frac{1}{\Gamma(b)} \cdot \frac{\Gamma(b)}{(1 - t)^b} = \frac{1}{(1 - t)^b} \quad t < 1  \\
            \mathcal{L}\{p(\gamma)\}(t) &= (1 + t)^{-b}
    \end{aligned}
\end{equation}

\subsection{Understanding Polya-Gamma distribution}
The Polya-Gamma distribution is a class of distributions that are used to sample from the posterior distribution of the logistic regression model.
Specifically, the random variables $w$ drawn from the Polya-Gamma distribution is equivalent to the infinite sum of random variables drawn from gamma distribution with shape parameter $b$ and scale parameter $1$.
In more detail, the Polya-Gamma distribution is defined as follows:

\begin{equation}
    \begin{aligned}
    \label{eq_polya_gamma}
    w \sim \text{PG}(b, c) &\leftarrow w = \frac{1}{2 \pi^2} \sum_{k=1}^{\infty} \frac{\gamma_k}{\left(k  - \frac{1}{2}\right)^2 + \left(\frac{c}{4 \pi^2}\right)^2}  \\
                            & \text{where} \quad \gamma_k \sim \text{Gamma}(b, 1), \quad p(\gamma_k) = \frac{1}{\Gamma(b)} \gamma^{b-1} e^{-\gamma} \\
    \end{aligned}
\end{equation}

We consider the laplace transform of $\omega \sim$ polya-gamma distribution (b, 0), we have:
\begin{equation}
    \begin{aligned}
    \label{eq_laplace_gamma}
        \mathbb{E}\left[ e^{-t \omega} \right] &= \mathbb{E}\left[e^{-t \frac{1}{2 \pi^2} \sum_{k=1}^{\infty} \frac{\gamma_k}{\left(k  - \frac{1}{2}\right)^2}} \right] \\
                                               &= \prod_{k=1}^{\infty} \mathbb{E} \left[ e^{\frac{-t}{2 \pi^2} \frac{\gamma_k}{\left(k  - \frac{1}{2}\right)^2}}\right]
    \end{aligned}
\end{equation}
where, $\gamma_k \sim \text{Gamma}(b, 0)$. We can simplify the expectation of the exponential term as follows
$\mathbb{E} \left[ e^{\frac{-t}{2 \pi^2} \frac{\gamma_k}{\left(k  - \frac{1}{2}\right)^2}}\right] = (1 + \frac{t}{2 \pi^2} \frac{1}{\left(k  - \frac{1}{2}\right)^2})^{-b}$ by $\mathbb{E} \left[ e^{-t \gamma_k}  \right] = (1 + t)^{-b}$.
By organizing the results, we substitute back to the Laplace transform of Polya-Gamma distribution:
\begin{equation}
    \begin{aligned}
    \label{eq_laplace_pg_1}
        \mathbb{E}\left[ e^{-t \omega} \right] &= \prod_{k=1}^{\infty} \left( 1 + \frac{t}{2 \pi^2 \left(k  - \frac{1}{2}\right)^2} \right)^{-b} \\
    \end{aligned}
\end{equation}
Next, we consider the $\cosh(x)$ from \autoref{eq_coshx_infinite_product}. It is directly related to the Laplace transform of Polya-Gamma distribution. 
Specifically, the equation $\cosh(\frac{\sqrt{t}}{2}) = \prod_{k=1}^{\infty} \left( 1 + \frac{t}{2 \pi^2 \left(k  - \frac{1}{2}\right)^2} \right)$ allows to rewrite the equation as follows. 
\begin{equation}
    \begin{aligned}
    \label{eq_laplace_pg_2}
        \mathbb{E}\left[ e^{-t \omega} \right] = \left( \cosh(\frac{\sqrt{t}}{2}) \right)^{-b} \\
    \end{aligned}
\end{equation}
Hence, we obseve that the product form of $\cosh(x)$ leads to the expression of the Laplace transform of Polya-Gamma distribution.

\subsection{Revisiting Sigmoid Function}
We consider the general form of the sigmoid function \autoref{eq_sigmoid_function} as $ \frac{\left( e^{f} \right)^\alpha}{\left( 1+e^f \right)^b}$.
First, we dervie $\left(1 + e^{f} \right)^{-b} = 2e^{\frac{f}{2}}\cosh \left( \frac{f}{2} \right)^{-b}$. 
We know $\cosh \left( \frac{x}{2} \right) = \frac{e^{\nicefrac{x}{2}} - e^{-\nicefrac{x}{2}}}{2}$ by \autoref{eq_zero_cosh}. 
$(1+e^{f}) = e^{f/2} \left( e^{f/2} + e^{-f/2} \right)$ allows to rewrite as follow:
\begin{equation}
    \centering
    \begin{aligned}
    \label{eq_denominator_sigmoid}
        \left( 1+e^{f} \right) = e^{f/2} \left( e^{f/2} + e^{-f/2} \right) = 2e^{f/2} \cosh \left( \frac{f}{2} \right) \\
        \left( 1+e^{f} \right)^{-b} = 2^{-b}e^{-bf/2} \left( \cosh \left( \frac{f}{2} \right) \right)^{-b} \\
    \end{aligned}
\end{equation}
We observe that $\cosh \left( \frac{\sqrt{t}}{2} \right)$ is equivalent to MGF of Polya-Gamma distribution $\mathbb{E}_{\omega} \left[ e^{-t \omega} \right]$.
By substituting $t = \nicefrac{\phi^2}{2}$, we can write the following relationship:
\begin{equation}
    \centering
    \begin{aligned}
    \label{eq_denominator_sigmoid_2}
        \left(\cosh \left( \frac{\sqrt{t}}{2} \right) \right)^{-b} &=  \int_{0}^{\infty} e^{-t \omega} p(\omega) d\omega \quad \text{where,} \quad p(\omega) \sim \text(PG)(b, 0) \\
        \left( \cosh \left( \frac{\phi}{2} \right) \right)^{-b} &= \int_{0}^{\infty} e^{-\frac{\phi^2}{2} \cdot \omega} p(\omega) d\omega
    \end{aligned}
\end{equation}
This equation $(1+e^{f})^{-b} = 2^{-b}e^{-bf/2} \int_{0}^{\infty} e^{-\nicefrac{\phi^2}{2} \cdot \omega} p(\omega) d\omega$ allows us to rewrite the sigmoid function in more general way. Specifically, we have
\begin{equation}
    \centering
    \begin{aligned}
    \label{eq_general_sigmoid}
        \frac{(e^{f})^{\alpha}}{(1+e^{f})^{b}} &= 2^{-b}{e^{f}}^{\alpha}e^{-bf/2} \int_{0}^{\infty} e^{-\nicefrac{\phi^2}{2} \cdot \omega} p(\omega) d\omega  \\
        \frac{(e^{f})^{\alpha}}{(1+e^{f})^{b}} &= 2^{-b}e^{f (\alpha - \nicefrac{b}{2})} \int_{0}^{\infty} e^{-\nicefrac{\phi^2}{2} \cdot \omega} p(\omega) d\omega  \\
    \end{aligned}
\end{equation}
We substitute $\kappa = \alpha - \nicefrac{b}{2}$ and allow as follows.
\begin{equation}
    \centering
    \begin{aligned}
    \label{eq_final_general_sigmoid}
    \frac{(e^{f})^{\alpha}}{(1+e^{f})^{b}} &= 2^{-b}e^{f \kappa} \int_{0}^{\infty} e^{-\nicefrac{\phi^2}{2} \cdot \omega} p(\omega) d\omega  \\
    \end{aligned}
\end{equation}

Suppose to have $\alpha_i = y_i, b = 1, \kappa_i = y_i - \nicefrac{1}{2}$ and $f_i = \vbeta^T\vx_i$, we can rewrite the sigmoid function $\sigma(f) = \frac{1}{1+e^{-f_i}}$ as follows:
\begin{equation}
    \centering
    \begin{aligned}
    \label{eq_example_sigmoid_as_pg}
        p(y_i|f_i) &= \sigma(f_i)^{y_i} (1-\sigma(f_i))^{1-y_i} = \left( \frac{e^{f_i}}{1+e^{f_i}} \right)^{y_i} \left( \frac{1}{1+e^{f_i}} \right)^{1-y_i} \\ 
        &= \frac{e^{{f_i}^{y_i}}}{1+e^{f_i}} = \frac{1}{2} e^{\kappa_i f_i} \int_{0}^{\infty} e^{-\frac{\omega_i f_i^2}{2}} p(\omega_i) d\omega_i\\
    \end{aligned}
\end{equation}
In this context, we view the likelihood $p(y_i|f_i)=\frac{e^{{f_i}^{y_i}}}{1+e^{f_i}} = \frac{1}{2} e^{\kappa_i f_i} \int_{0}^{\infty} e^{-\frac{\omega_i f_i^2}{2}} p(\omega_i) d\omega_i$ as an integral over the polya-gamma distribution,
where we interpret this expression as $\int p(y_i, \omega_i|f_i) d\omega_i$.
By this interpretation, the joint distribution $p(y_i, \omega_i|f_i)$ enables us to find the conditional distributions $p(y_i|\omega_i, f_i) = \frac{p(y_i, \omega_i|f_i)}{p(\omega_i|f_i)} = \frac{p(y_i, \omega_i|f_i)}{p(\omega_i)} \propto p(y_i, \omega_i|f_i)$.

When we consider the relationship $p(y_i|\omega_i, f_i) \propto \frac{e^{{f_i}^{y_i}}}{1+e^{f_i}} = \frac{1}{2} e^{\kappa_i f_i} \int_{0}^{\infty} e^{-\frac{\omega_i f_i^2}{2}} p(\omega_i) d\omega_i$, 
it implies that $p(y_i|\omega_i, f_i)$ is a gaussian distribution over $f$. 
Given that $\omega_i$ is drawn from polya-gamma distribution as shown in \autoref{eq_polya_gamma}, 
the conditional distribution of $y_i$ given $f_i$ and $\omega_i$ simplifies to $p(y_i|f_i, \omega_i) \propto e^{-\frac{\omega_i {f_i}^2}{2}} e^{\kappa_i f_i}$.
The detail derivation proceeds as follows.
\begin{equation}
    \begin{aligned}
        \label{eq_y_approx_normal_dist}
        p(y_i \mid f_i, \omega_i) &\propto e^{-\frac{\omega_i {f_i}^2}{2}} e^{\kappa_i f_i} 
        = e^{-\frac{\omega_i {f_i}^2}{2} + \kappa_i f_i} \\
        & = e^{ -\frac{\omega_i}{2}\left(f_i^2 - \frac{2\kappa_i}{\omega_i}f_i + \frac{\kappa_i^2}{\omega_i^2} - \frac{\kappa_i^2}{\omega_i^2} \right)} \\
        & = e^{ -\frac{\omega_i}{2}\left(f_i - \frac{\kappa_i}{\omega_i} \right)^2 + \frac{\omega_i}{2}\frac{\kappa_i^2}{\omega_i^2}} \\
        & = e^{ -\frac{\omega_i}{2}\left(f_i - \frac{\kappa_i}{\omega_i} \right)^2} e^{\frac{\kappa_i^2}{2\omega_i}} \\
        p(y_i \mid f_i, \omega_i) &\propto e^{ -\frac{\omega_i}{2}\left(f_i - \frac{\kappa_i}{\omega_i} \right)^2}
    \end{aligned}
\end{equation}
It implies that the conditional distribution of $y_i$ given $f_i$ and $\omega_i$ is a gaussian distribution $\mathcal{N}(\frac{\kappa_i}{\omega_i}; f_i, \omega_i^{-1})$.
For convenience, let us introduce a new variable $z_i$ such that $\frac{\kappa_i}{\omega_i}$ and it is related to the label $y_i$.
The multivariate cases describe the conditional distribution $\mathcal{N}(\boldsymbol{\Omega}^{-1}\vkappa_i; \textbf{f}_i, \boldsymbol{\Omega}^{-1})$ where $\vkappa_i, \vf_i \in \mathbb{R}^{\mathcal{C}}$ and $\boldsymbol{\Omega} = \omega_i \mI$.

\subsection{Construction of Conjugate Relationship for Posterior Distribution $f_i$ and $\omega_i$.}
Prior to derviation of posterior distribution, we cover the construction of conjugate relationship for gaussian distribution. 

\begin{definition}
    \label{eq_marginal_dist}
    Given a margnial gaussian distribution of $\textbf{b}$ and a conditional gaussian distribution for $\textbf{a}$ given $\textbf{b}$,
    We define these gaussian distributions as follows:

    \begin{equation}
        \begin{aligned}
            \label{eq_gassian_distributions}
            p(\textbf{b}) &= \mathcal{N}(\textbf{b}; \textbf{m}, \textbf{S}^{-1}) \\
            p(\textbf{a} \mid \textbf{b}) &= \mathcal{N}(\textbf{a}; \textbf{W} \textbf{b}, \boldsymbol{\Phi}^{-1}) \\
        \end{aligned}
    \end{equation}
\end{definition}

\begin{lemma}
    \label{eq_induced_dist}
    The marginal distribution of $\textbf{a}$ and the conditional distribution of $\textbf{b}$ given $\textbf{a}$ are given by    
    \begin{equation}
        \begin{aligned}
            \label{eq_induced_gaussian_distributions}
            p(\textbf{b} \mid \textbf{a}) &= \mathcal{N}(\textbf{b}; \boldsymbol{\mu}, \boldsymbol{\Sigma}) \\
            p(\textbf{a}) &= \mathcal{N}(\textbf{a}; \textbf{W} \textbf{m}, \boldsymbol{\Phi}^{-1} + \textbf{W} \textbf{S}^{-1} \textbf{W}^{T}) \\
        \end{aligned}
    \end{equation}
    where $\boldsymbol{\Sigma} = (\textbf{S} + \textbf{W}^{T} \boldsymbol{\Phi} \textbf{W})^{-1}$ and $\boldsymbol{\mu} = \boldsymbol{\Sigma} (\textbf{W}^{T} \boldsymbol{\Phi} \textbf{a} + \textbf{S} \textbf{m})$.
\end{lemma}

\subsubsection{Posterior Distribution $f_i$}
Suppose that $f_i = \boldsymbol{\beta}^T\vx_i$ in \autoref{eq_sigmoid_function} is parameterized by $\vbeta$, 
we can define the prior and posterior distribution for $\vbeta$ instead of $f_i$. 
In this case, we consider the prior distribution $p(\boldsymbol{\beta})$ as $ \boldsymbol{\beta} \sim \mathcal{N}(\boldsymbol{\beta}; \boldsymbol{m}, \boldsymbol{S}^{-1})$. 
The posterior distribution $p(\boldsymbol{\beta}|y_i, \omega_i)$ can be defined by \autoref{eq_induced_gaussian_distributions} with $p(y_i, |f_i, \omega_i)$ and $p(\boldsymbol{\beta})$.
It can be described as $p(\boldsymbol{\beta}|y_i, \omega_i) = \mathcal{N}(\boldsymbol{\beta}; \boldsymbol{\mu}, \boldsymbol{\Sigma})$ 
where $\boldsymbol{\Sigma} = (\textbf{S} + \boldsymbol{\Omega})^{-1}$ 
and $\boldsymbol{\mu} = \boldsymbol{\Sigma} (\boldsymbol{\kappa} + \textbf{S} \textbf{m})$.

\begin{equation}
    \begin{aligned}
        \label{eq_posterior_beta}
        p(f_i \mid y_i, \omega_i) &= p(\boldsymbol{\beta} \mid y_i, \omega_i) = \mathcal{N}(\boldsymbol{\beta}; \boldsymbol{\mu}, \boldsymbol{\Sigma}) \\
        \text{where} \quad \boldsymbol{\mu} &= \boldsymbol{\Sigma} (\boldsymbol{\kappa} + \textbf{S} \textbf{m}) \\
        \boldsymbol{\Sigma} &= (\textbf{S} + \boldsymbol{\Omega})^{-1}
    \end{aligned}
\end{equation}

Next, we derive the posterior distribution of $\omega_i$ given $y_i$ and $f_i$.
We start by considering the joint distribution $p(y_i, \omega_i | f_i)$.
We substitue $p(\omega_i | y_i, f_i)$ with $\frac{p(y_i, \omega_i|f_i)}{p(y_i|f_i)} = \frac{p(y_i|\omega_i, f_i)p(\omega_i|f_i)}{p(y_i|f_i)} = \frac{p(y_i|\omega_i, f_i)p(\omega_i)}{p(y_i|f_i)}$ due to the independence of $f_i$ and $\omega_i$.
Accordingly, we have $p(\omega_i | y_i, f_i) \propto p(y_i|\omega_i, f_i)p(\omega_i)$. 
By \autoref{eq_y_approx_normal_dist}, $p(y_i|\omega_i, f_i) = e^{-\frac{\omega_i {f_i}^2}{2}} e^{\kappa_i f_i}$ and $p(\omega_i) = p(\omega_i) \sim \text{PG}(1, 0)$ by definition.
\section{Detailed Illustration of Proposed Method}
The procedure of the proposed method is outlined through its mathematical framework, beginning with tensor operations related to $A$ and $f$. The next part explains the One-vs-Each SoftMax approximation. The final subsection describes the conjugacy of the One-vs-Each SoftMax approximation. Each component of the sigmoid function in this approximation shows conjugacy with Polya-Gamma variables, which are obtained from pre-trained models. Additionally, the label information from the training data helps define the posterior distribution, allowing the fine-tuned model to learn as expected. 

\subsection{Einstein Notation and Tensor Operations}
\label{app_einstein_op}
We define the tensor $A$ using Kronecker delta functions. Here, $A \in \mathbb{R}^{C \times C \times C}$ captures the differences between identity matrices along specified dimensions.
\begin{equation}
    \begin{aligned}
    \label{eq_affine_matrix_2}
        A_{ijk} = \delta_{ik} - \delta_{jk}, \quad \text{for } i, j, k = 1, \dots, C
    \end{aligned}
\end{equation}

Next, we consider a matrix $f$  of size  $N \times C$. Here, $N$  represents the number of samples, and  $C$  is the number of classes.
\begin{equation}
    \begin{aligned}
    \label{eq_logit_function}
        f \in \mathbb{R}^{N \times C} =
        \begin{bmatrix}
        f_{11} & \dots & f_{1C} \\
        f_{21} & \dots & f_{2C} \\
        \vdots & \ddots & \vdots \\
        f_{N1} & \dots & f_{NC}
        \end{bmatrix}
    \end{aligned}
\end{equation}

We aim to compute the tensor product $Af$, defined as:
\begin{equation}
    \begin{aligned}
    \label{eq_tensor_product}
        \psi_{nij} = (Af)_{nij} = f_{ni} - f_{nj}, \quad \text{for} \quad n = 1, \dots, N; \; i, j = 1, \dots, C
    \end{aligned}
\end{equation}

This operation results in a tensor $Af \in \mathbb{R}^{N \times C \times C}$, where each element represents the difference between elements of $f$ across different classes for each sample.
To illustrate this explicitly, we can write out the tensor  $Af$  for each sample $n$:

\begin{equation}
    \begin{aligned}
    \label{eq_explicit_form}
        \psi = Af =
        \begin{bmatrix}
        \begin{bmatrix}
        0 & f_{n1} - f_{n2} & \dots & f_{n1} - f_{nC} \\
        f_{n2} - f_{n1} & 0 & \dots & f_{n2} - f_{nC} \\
        \vdots & \vdots & \ddots & \vdots \\
        f_{nC} - f_{n1} & f_{nC} - f_{n2} & \dots & 0
        \end{bmatrix}
        \Bigg|_{n=1}^{N}
        \end{bmatrix}
    \end{aligned}
\end{equation}

This explicit form shows that for each sample $n$, the tensor $Af$ contains all pairwise differences of the elements of $f$ for that sample.
By representing the operation in this way, we clarify how the tensor product operates between $A$ and $f$. This formulation can be efficiently executed using tensor manipulation functions such as \texttt{torch.einsum} in PyTorch, which is well-suited for handling complex tensor operations. To further illustrate this, the following subsection provides a straightforward example involving three-class classification, showcasing the practical application of this concept.

\subsection{Simple Example: One-vs-Each SoftMax Approximation on Three-Class Classification}
\label{app_one_vs_each_calculation}
Let us now provide the tensor product based on Einstein summation convention $\mA \vf$ in a practical implementation. Consider a classification task with three classes, where $C = 3$. The tensor $A \in \mathbb{R}^{3 \times 3 \times 3}$ is defined as
\begin{equation}
\begin{aligned}
\label{eq_affine_matrix_ex}
A \in \mathbb{R}^{C \times C \times C} &= 
\begin{bmatrix}
    \begin{bmatrix}
        0 & 0 & 0 \\
        1 & -1 & 0 \\
        1 & 0 & -1
    \end{bmatrix} \\[2em]
    \begin{bmatrix}
        -1 & 1 & 0 \\
        0 & 0 & 0 \\
        0 & 1 & -1
    \end{bmatrix} \\[2em]
    \begin{bmatrix}
        -1 & 0 & 1 \\
        0 & -1 & 1 \\
        0 & 0 & 0 \\
    \end{bmatrix}
\end{bmatrix}
\end{aligned}
\end{equation}

The original logit matrix $\vf \in \mathbb{R}^{N \times 3}$ contains the logit values for $N$ samples and 3 classes
\begin{equation}
\begin{aligned}
\label{eq_logit_matrix_ex}
\vf \in \mathbb{R}^{N \times C} &= 
\begin{bmatrix}
    \vf_1 \\
    \vf_2 \\
    \vdots \\
    \vf_N
\end{bmatrix}
=
\begin{bmatrix}
    f_{11} & \cdots & f_{1C} \\
    f_{21} & \cdots & f_{2C} \\
    \vdots & \ddots & \vdots \\
    f_{N1} & \cdots & f_{NC}
\end{bmatrix}
\end{aligned}
\end{equation}

The Einstein summation notation $(\mA \vf)_{nic} = \sum_j \mA_{ijc} \vf_{ni}$, denoted by $\vpsi$, computes the pairwise differences between the feature values across classes for each sample

\begin{equation}
\begin{aligned}
\label{eq_AF_ex}
\vpsi = \mA \vf &= 
\begin{bmatrix}
    \begin{bmatrix}
        0 & f_{11} - f_{12} & f_{11} - f_{13} \\
        f_{12} - f_{11} & 0 & f_{12} - f_{13} \\
        f_{13} - f_{11} & f_{13} - f_{12} & 0
    \end{bmatrix} \quad (n=1) \\[2em]
    \begin{bmatrix}
        0 & f_{21} - f_{22} & f_{21} - f_{23} \\
        f_{22} - f_{21} & 0 & f_{22} - f_{23} \\
        f_{23} - f_{21} & f_{23} - f_{22} & 0
    \end{bmatrix} \quad (n=2) \\[2em]
    \vdots \\[2em]
    \begin{bmatrix}
        0 & f_{n1} - f_{n2} & f_{n1} - f_{n3} \\
        f_{n2} - f_{n1} & 0 & f_{n2} - f_{n3} \\
        f_{n3} - f_{n1} & f_{n3} - f_{n2} & 0
    \end{bmatrix} \quad (n=N)
\end{bmatrix}
\end{aligned}
\end{equation}

This example makes it clear how the tensor operations introduced in the previous subsection apply to the implementation scenario. Each matrix $\psi_n$ for sample $n$ contains the pairwise differences between the class logit values. Next, we compute the One-vs-Each SoftMax approximation using a sigmoid function $\sigma(\psi_{nij})$ and apply it for classification. 

\begin{small}
\begin{equation}
\begin{aligned}
\label{eq_psi_ex}
    \prod_{i=1}^{3} \sigma\left(\psi_{nic}\right) &= \prod_{j=1}^{3} \sigma \left( \left(\mA \vf  \right)_{nic} \right) \in \mathbb{R}^{n \times C} \\
    &= \begin{pmatrix}
        \sigma(0) \sigma(\psi_{112}) \sigma(\psi_{113}) & \sigma(\psi_{121}) \sigma(0) \sigma(\psi_{123}) & \sigma(\psi_{131}) \sigma(\psi_{132}) \sigma(0) \\
        \sigma(0) \sigma(\psi_{212}) \sigma(\psi_{213}) & \sigma(\psi_{221}) \sigma(0) \sigma(\psi_{223}) & \sigma(\psi_{231}) \sigma(\psi_{232}) \sigma(0) \\
        \vdots & \vdots & \vdots \\
        \sigma(0) \sigma(\psi_{n12}) \sigma(\psi_{n13}) & \sigma(\psi_{n21}) \sigma(0) \sigma(\psi_{n23}) & \sigma(\psi_{n31}) \sigma(\psi_{n32}) \sigma(0)
    \end{pmatrix} \\
    \vspace{0.1in} \\
    &= \begin{pmatrix}
        \sigma(0) \sigma(f_{11} - f_{12}) \sigma(f_{11} - f_{13}) & \sigma(f_{12} - f_{11}) \sigma(0) \sigma(f_{12} - f_{13}) & \sigma(f_{13} - f_{11}) \sigma(f_{13} - f_{12}) \sigma(0) \\
        \sigma(0) \sigma(f_{21} - f_{22}) \sigma(f_{21} - f_{23}) & \sigma(f_{22} - f_{21}) \sigma(0) \sigma(f_{22} - f_{23}) & \sigma(f_{23} - f_{21}) \sigma(f_{23} - f_{22}) \sigma(0) \\
        \vdots & \vdots & \vdots \\
        \sigma(0) \sigma(f_{n1} - f_{n2}) \sigma(f_{n1} - f_{n3}) & \sigma(f_{n2} - f_{n1}) \sigma(0) \sigma(f_{n2} - f_{n3}) & \sigma(f_{n3} - f_{n1}) \sigma(f_{n3} - f_{n2}) \sigma(0)
    \end{pmatrix}
\end{aligned}
\end{equation}
\end{small}

The final output is denoted as $ \vf_{\theta}^{m}$ in Algorithm \ref{algo:pg_learning}, which is used in the loss function

\begin{equation}
\begin{aligned}
\label{eq_psi_revisited_ex}
    \vf_{\theta}^{m} &= \prod_{i=1}^{3} \sigma\left(\psi_{nic}\right) \in \mathbb{R}^{n \times C} \\
    &= \begin{pmatrix}
        \sigma(0) \sigma(\psi_{112}) \sigma(\psi_{113}) & \sigma(\psi_{121}) \sigma(0) \sigma(\psi_{123}) & \sigma(\psi_{131}) \sigma(\psi_{132}) \sigma(0) \\
        \sigma(0) \sigma(\psi_{212}) \sigma(\psi_{213}) & \sigma(\psi_{221}) \sigma(0) \sigma(\psi_{223}) & \sigma(\psi_{231}) \sigma(\psi_{232}) \sigma(0) \\
        \vdots & \vdots & \vdots \\
        \sigma(0) \sigma(\psi_{n12}) \sigma(\psi_{n13}) & \sigma(\psi_{n21}) \sigma(0) \sigma(\psi_{n23}) & \sigma(\psi_{n31}) \sigma(\psi_{n32}) \sigma(0)
    \end{pmatrix} \\
\end{aligned}
\end{equation}

In this example, we use the negative log-likelihood loss $\mathcal{L}_{\text{nll}}^{m} = \sum_{n} \sum_{c} y_{n}^c \log \left( {f_{\theta}}_n^c \right)^m$ for classification.
This example serves as a practical illustration of how the tensor operations introduced in the previous subsection can be used in implementation.

\subsection{One-vs-Each SoftMax Approximation with Polya-Gamma Augmentation}

Following the previous sections \autoref{app_polya_gamma} and \autoref{app_one_vs_each_calculation}, we now extend the One-vs-Each SoftMax approximation by introducing Polya-Gamma augmentation. Prior to explanation of this subsection, we implemente sampling procedure of posterior distribution in the $\psi$ space rather than $\vf$ space, because the matrix $A$ must be considered when working in the $\vf$ space shown in the main manuscript. For computational efficiency, sampling on the space of $\psi$ can minimize the computation burden based on our assumption. It makes the proposed method more feasible in terms of implementation. 
To maintain consistency with notation used in Algorithm \ref{algo:pg_learning}, we substitute $f$ from the previous subsection with $\mu$. Specifically, $\vf \sim \mathcal{N}(\vmu, \frac{1}{\alpha} \textbf{I})$ and $\vf_{\theta} \sim \mathcal{N}(\vmu_{\theta}, \frac{1}{\alpha} \textbf{I})$. 

This extension demonstrates that each sigmoid function in the tensor $\vpsi_{\theta} =  \mA \vmu_{\theta} \in \mathbb{R}^{n \times C \times C}$ (defined in \autoref{eq_AF_ex}) has its own conjugacy properties. Specifically, $\vpsi_{\theta} \sim \mathcal{N}(\mA \vmu_{\theta}, \nicefrac{2}{\alpha} \textbf{I})$. This arises because $A$ represents the subtract between $f_i$ and $f_j$ and since these terms are independent, the variances for $\psi_{\theta}$ are effectively doubled, leading to the simplified covaraince structure.
First, we define the conjugacy of the Polya-Gamma variables. The variable $\vkappa \in \mathbb{R}^{n \times C \times C}$ is derived from the one-hot encoded label information $\vy \in \mathbb{R}^{n \times C}$, along with the tensor $A$ described in \autoref{app_einstein_op}. These variables form the basis of the posterior distribution. The Polya-Gamma variable $\vomega \in \mathbb{R}^{n \times C \times C}$ is sampled using $\vpsi$ constructed by the pre-trained model $\vmu$ and tensor $A$. More Specifically, $\vpsi_{\theta}$ represents the fine-tuned model, while $\vpsi$ denotes the pre-trained model, both utilizing the same $A$.
We assume that each dimension of $\vpsi$, $\vkappa$, and $\vomega$ is independent, which simplifies the covariance structure for the posterior distribution. Under this assumption, the posterior covariance becomes a diagonal matrix, significantly reducing computational complexity.
The Polya-Gamma variable $\vomega$ is drawn as follows:
\begin{equation}
\begin{aligned}
\label{eq_omega_drawn}
    \vomega \sim \text{PG}(1, \vpsi) \in \mathbb{R}^{n \times C \times C}
\end{aligned}
\end{equation}
Using this formulation, we can update $\vpsi_{\theta}$ based on $\vkappa$ and $\vpsi_{\theta} = A\vmu_{\theta}$. Note that $\vpsi_{\theta}$ and $\vmu_{\theta}$ are distinct, as $\vmu_{\theta}$ indicates the original logit value induced by the fine-tuning model.
The posterior distribution for $\vpsi_{\theta}$ can be expressed as:
\begin{equation}
\begin{aligned}
  p(\vpsi_{\theta}^{(m)} | \vy, \vomega) &\propto p(\vy|\vpsi_{\theta}, \vomega) p(\vpsi_{\theta}), \nonumber \\
  &\propto \mathcal{N}(\vpsi_{\theta} | \boldsymbol{\tilde{\Sigma}} (\boldsymbol{\Sigma^{-1}} \vmu_{\theta}\mA + \vkappa), \boldsymbol{\tilde{\Sigma}})
\end{aligned}
\end{equation}
Here, $\tilde{\Sigma}$ is the posterior covariance, defined as:
\begin{equation}
\begin{aligned}
\label{app_eq_psi_theta}
\boldsymbol{\tilde{\Sigma}} = (\frac{\alpha}{2} \textbf{I} + \boldsymbol{\Omega})^{-1}
\end{aligned}
\end{equation}
where $\nicefrac{\alpha}{2} \textbf{I} \in \mathbb{R}^{ (n \times C \times C ) \times (n \times C \times C)}$ represents the identity matrix scaled by $\alpha$, and $\boldsymbol{\Omega} \in \mathbb{R}^{ (n \times C \times C ) \times (n \times C \times C)}$ is the Polya-Gamma covariance matrix. Since we assume that all components of the matrices are diagonal, computing the inverse of $\boldsymbol{\tilde{\Sigma}}$ is straightforward, as it reduces to taking the reciprocal of the diagonal components.
The drawn samples $\vpsi_{\theta}^{(m)} \in \mathbb{R}^{n \times C \times C}$ are fed into the sigmoid function as described in \autoref{eq_psi_ex}, ensuring One-vs-Each SoftMax approximation with polya-gamma augmentation.

The interpretation of this procedure is that the parameter $\vomega$, influenced by the pre-trained models, regulates the degree of confidence each sigmoid function derives from the pre-trained models. Additionally, the label information $\vkappa$ contributes to the adjustment of the posterior distribution, although it does not entirely determine the final logit values. This process ensures that the fine-tuned models incorporate both pre-trained knowledge and label information, rather than relying exclusively on label information. When updating the loss function $\mathcal{L}_{\text{nll}}$ with $\vpsi_{\theta}$, the parameter $\vpsi_{\theta}^{(m)}$ is influenced by both the label information and the pre-trained model. However, no single component completely dominates the process. This balance helps prevent overfitting, allowing the fine-tuned model $\psi_{\theta}$ to capture the general patterns of the training datasets without being overly influenced by individual data points.

\subsection{More Details on \autoref{algo:pg_learning}}
\label{subsec_app_details}

Here we would like to provide more details on our method. 

\subsubsection{Assumption on $\textbf{A} \approx \textbf{I}$}
As described in \autoref{subsec:param_est}, we assume $\mA \approx \mI$.
This means we can write down the posterior over $\psi$ for each class separately, and they coincide. Hence, the posterior over $\vpsi$ for class $c$ is given by 
   \begin{align}
p(\vpsi^c| \vomega^c, \Dat)
     &\propto \mathcal{N}(\vpsi^c | \bm{\Sigma}^c(\frac{\alpha}{2} \mA^{c}\bm{\mu}^c + \bm{\kappa}^c), \bm{\Sigma}^c)
\end{align} where $\bm{\Sigma}^c = (\frac{\alpha}{2} \mI_n + \bm\Omega^c)^{-1}$.
This construction is motivated by purely computational reasons. By assuming the functions across classes are independent, we avoid computing the large covariance matrix of $(n\times C\times C) \times (n \times C \times C)$, which is prohibitive when $C$ is large. The size of the covariances in each class is $(n \times C) \times (n \times C)$, which is manageable when $n$ is small, as in our case.  


\subsubsection{Two Different Priors}

In practice, we use $\vf \sim \Nrm(\vmu, \alpha^{-1} \mI)$ as a prior when we do the posterior inference on the PG variables $\vomega$. On the other hand, we use $\vf_\vtheta \sim \Nrm(\vmu_\vtheta, \alpha^{-1} \mI)$ as a prior when we do the posterior inference on the function $\vf_\vtheta$ (Sampling of $\vpsi^{(m)}_\vtheta$ in \autoref{algo:pg_learning} indicates that). 

We acknowledge that this formulation deviates from the usual Bayesian construction. 
Our reasoning for this construction is that this simplifies the posterior updates over $\vomega$ and $\vf_\vtheta$. In the usual case where there are no parameters to optimize, we employ the Gibs steps, which involve alternating the updates for $\vomega_\vtheta$ given $\vf_\vtheta$  and $\vf_\vtheta$ given $\vomega_\vtheta$. This is computationally costly because whenever we step toward new values of $\vtheta$ in optimization, we need to alternate these two steps. By setting $\vomega$ as a function of the prior $\vf$ (i.e., parameter-independent), we do not update $\vomega$, resulting in speedy optimization.   

\subsubsection{Rigorous Definition of the Posterior over $\textbf{f}$}

We approximate the posterior over $\vf_\vtheta$ defined as \autoref{eq:posterior_f_GP} using \autoref{eq:correct_post_f}.
In this case, the rigorous KL term yields $\| (\alpha \mI + \mA\trp\Omega \mA)^{-1}(\alpha \vmu_\vtheta + \mA \vkappa) - \vmu \|_2^2$ which depends on the parameters $\vtheta$.
We approximately reduce this to  $\| c\vmu_\vtheta - \vmu \|_2^2$, where c is some constant with respect to the parameters $\vtheta$. So, we still maintain the same regularization term given in the manuscript. 

\subsection{Algorithm of Proposed Method for CoOp\_OVE\_PG}

In Algorithm \ref{algo:pg_learning}, this explanation focuses on the CoOp version. In this context, we compare the proposed method with the CoOp version. The primary distinction between CoOp and CoCoOp lies in their approach to prompt learning. Specifically, CoOp employs the expression $I(X)^T T(\vp_{\vtheta_1}^c)$, whereas CoCoOp utilizes $I(X)^T T(\vp_{\vtheta_1}^c + \vr_{\vtheta_2}^c(\vx))$. The detailed procedure for this is outlined below.

\begin{algorithm}[!ht]
  \caption{One-vs-Each SoftMax Approximation with P\'olya-Gamma Auxiliary Variables for Prompt Learning (CoOp based)}
  \label{algo:pg_learning_coop}
  \begin{algorithmic}
    \STATE {\bfseries Input:} Objective
    $\mathcal{L}_{\text{elbo}} = \{\mathcal{L}_{\text{nll}}, \mathcal{L}_{\text{KLD}} \}$, Class prompts $\mathcal{C} \in \{c_1, \cdots, c_C \}$, The learnable prompts $\textbf{p}_{\vtheta}$,
    The predefined prompts for the original CLIP $\textbf{t} = \{ t_1, \cdots, t_C \}$, Image-Encoder $I$ and Text-encoder $T$,
    The number of parallel Gibbs chains $M$, Total number of classes $C$. Learning rate $\eta$, Weight for KLD $\beta$, $\alpha$ prior precision of $\vf$, Einstein summation convention, denoted by $\vpsi = \mA \vf$.
    \vspace{.1cm} 
    
    \STATE {\bfseries Initialize:} Parameters $\vtheta$ randomly.
    \REPEAT \STATE Sample a mini-batch $(\vx, \vy)$ from a training data
    $\Dat, \text{where} \ \vx \in \mathbb{R}^{n \times d}, \vy \in \{0,1\}^{n \times C}$
    
    \STATE $\mA \leftarrow {\text{OVE-MATRIX}}(\bf{\vy}) \in \mathbb{R}^{C \times C \times C}$ in \autoref{eq:A}

    $\vmu_{\vtheta} = I(\vx)^T T(\vp_{\vtheta}^c) \in \mathbb{R}^{n \times C} $  
        
    $\vmu = I(\vx)^T T(\vp^c) \in \mathbb{R}^{n \times C} $

    \STATE $\vpsi_{\vtheta, nic} \leftarrow \sum_j \mA_{ijc} {\vmu_{\vtheta, ni}}, \quad \vpsi_{\vtheta} \in \mathbb{R}^{N \times C \times C}$
    \STATE $\vpsi_{nic} \leftarrow \sum_j \mA_{ijc} {\vmu}_{ni}, \quad \vpsi \in \mathbb{R}^{n \times C \times C} $  
    \STATE $\vkappa_{nic} = \sum_j \mA_{ijc} \big(\vy_{ni} - \nicefrac{1}{2}\big), \quad \vkappa \in \mathbb{R}^{n \times C \times C}$

    \FOR{$m=1$ {\bfseries to} $M$} 
    
        \STATE $\vomega^{(m)} \sim \text{PG}(1, \vpsi)$ \hfill given in \autoref{eq:posterior_omega_PG}     

        \STATE $\vpsi^{(m)}_{\vtheta} \sim p(\vpsi^{(m)} | \vomega^{(m)}, \vpsi_{\vtheta}, \vkappa, \alpha )$ \hfill given in \autoref{eq:posterior_f}

        \STATE $\vf^{(m)}_{\vtheta} \leftarrow \sum_C \vpsi^{(m)}_{\vtheta} \in \mathbb{R}^{n \times C}$ 
        \hfill due to $\vpsi = \bm{A} \vf_\vtheta$

        \STATE $\mathcal{L}_{\text{nll}}^{(m)} \leftarrow - \sum_{i=1}^{N} \sum_{c=1}^{C} y_{i,c} \log f_{i,c}$ \hfill given in \autoref{eq:OVE_likelihood} 

        \STATE $\mathcal{L}_{\text{KLD}}^{(m)} \leftarrow \beta \lVert \vmu_{\vtheta} - \vmu  \rVert_2^2$
    \ENDFOR 

    \STATE $\theta \leftarrow \theta - \frac{\eta}{M} \sum_{m=1}^M \nabla_{\bf{\theta}} \big( \mathcal{L}_{\text{nll}}^{m} + \mathcal{L}_{\text{KLD}}^{m} \big)$
    
    \UNTIL{convergence}
  \end{algorithmic}
\end{algorithm}

\clearpage

\section{Experiments for 1D Multi-Class Classification}
This section illustrates the experimental results of 3rd-order polynomial multi-class classification using SoftMax, OVE SoftMax approximation and the proposed method. The ground truth data is created by predefined Gaussian distributions. Three Gaussian distributions are used to represent the ground-truth classes: $p(x) \sim \mathcal{N}(1.0, 1.0^2)$,  $q(x) \sim \mathcal{N}(0.0, 2.0^2)$ and $r(x) \sim \mathcal{N}(-1.0, 1.0^2)$. These distributions have different means and variances to create distinguishable classes. The probabilistic densities of ground truth and the samples from each distribution are described in the first column of \autoref{fig:1D_classification}. The sample $X$ is transformed using a polynomial basis function $\Phi(X)$, which is of degree 3 in this case. The learnable parameters $\beta$ are applied to this transformed feature space. More specifically, $X \in \mathbb{R}^{N \times 1}$ where $N$ is the number of data points, and 1 is the dimensionality of the input data. $\Phi(X) \in \mathbb{R}^{N \times 3}$ where applying the basis function of degree 3. 

For this multi-class classification problem, which involves three classes, there are three sets of weights, each of size 3, represented by $\beta = (\beta_1, \beta_2, \beta_3) \in \mathbb{R}^{3 \times 3}$. The second column of \autoref{fig:1D_classification} displays the results where $\beta$ is trained using the standard SoftMax function. The third column shows the results when $\beta$ is trained using the OVE SoftMax approximation. The fourth and fifth columns present the results of the proposed method, where $\beta$ is trained with different values of the hyperparameter  $\alpha$, specifically  $\alpha = 1$  and  $\alpha = 1000$, respectively. The source code for these experiments is provided in the supplementary materials.

\section{Experiments for Fine-tuning on E-MNIST with Pre-trained ViT on MNIST}

This section presents the experimental results of fine-tuning a pre-trained Vision Transformer (ViT) on the E-MNIST dataset, where the model was initially pre-trained on MNIST. To demonstrate the generalization capabilities of the proposed method across different tasks, we provide a simple example. First, we describe the ViT architecture, where its parameters are adapted for both the MNIST and E-MNIST datasets. Specifically, we use the ViT-3/7 structure with  $D=64$  (embedding dimension) and  $H=4$ (number of heads).
During the pre-training phase, the model is trained on the full MNIST dataset. For the fine-tuning phase, we use a subset of the original E-MNIST Digit dataset, sampling only the first 100 data points from each class to simulate a typical fine-tuning scenario. Both MNIST and E-MNIST consist of  $28 \times 28$  grayscale images, and the class labels between the two datasets align.
For evaluation, we test the fine-tuned model on the E-MNIST Digit test dataset, which contains data points the model has not encountered during training. This setup ensures a rigorous evaluation of the model’s ability to generalize from the pre-trained task (MNIST) to the fine-tuning task (E-MNIST).
The source code for these experiments is included in the supplementary material.


\section{Extended Experimental Results}
Our implementation is based on the code from \citet{derakhshani2023bayesian}\footnote{\texttt{https://github.com/saic-fi/Bayesian-Prompt-Learning}}. All experiments were conducted on a single NVIDIA RTX 4090. Due to our limited computational resources, we conducted the experiments with a batch size of 4 for CoOp, MaPLe, APEX and 1 for CoCoOp. We followed the hyper-parameter setup from \citet{derakhshani2023bayesian}.

For the seen-to-unseen generalization experiments, the seen and unseen categories were split according to the original CoOp work \citep{zhou2022learning}, which is a common setup for seen and unseen tasks across a wide range of studies \citep{zhou2022conditional, khattakMaPLe, yang-etal-2024-towards}. While we reported the results for the unseen categories in our main manuscript, we further provide results for both seen and unseen categories, along with their harmonic means. All experiments were conducted with 3 random seeds, and we report the standard deviations of these trials in \autoref{table:coop_all} and \autoref{table:cocoop_all}. 

\begin{table*}[t]
    \centering
    \resizebox{0.97\linewidth}{!}{
    \begin{tabular}{l|cc|cc|cc|cc}
    \toprule[0.1em]
                        & \multicolumn{2}{c|}{CoOp} & \multicolumn{2}{c|}{CoCoOp} & \multicolumn{2}{c|}{MaPLe} & \multicolumn{2}{c}{APEX} \\
                        &   SoftMax    &  OVE-PG   &   Softmax   &   OVE-PG  &   SoftMax    &  OVE-PG &   SoftMax    &  OVE-PG \\
        \midrule
        Caltech101      & 97.70 & 97.87 & 97.33 & 97.33 & 97.85 & 97.67 & 97.77 & 97.53 \\
        DTD             & 79.37 & 77.43 & 74.30 & 75.17 & 80.13 & 81.97 & 78.73 & 81.37 \\
        EuroSAT         & 87.77 & 85.27 & 84.83 & 84.73 & 89.67 & 90.43 & 90.03 & 84.63 \\
        FGVC Aircraft   & 33.60 & 32.73 & 35.83 & 35.79 & 38.67 & 36.60 & 36.77 & 35.20 \\
        Food101         & 90.43 & 90.60 & 90.67 & 90.43 & 90.57 & 90.37 & 90.70 & 90.40 \\
        Flowers102      & 95.27 & 92.60 & 92.53 & 92.20 & 96.00 & 95.70 & 95.10 & 94.73 \\
        Oxford Pet      & 95.00 & 95.13 & 95.33 & 94.03 & 95.60 & 95.30 & 95.40 & 95.13 \\
        Stanford Cars   & 72.30 & 67.37 & 69.80 & 69.83 & 74.00 & 70.17 & 73.17 & 70.93 \\
        SUN397          & 80.90 & 78.27 & 79.33 & 76.40 & 81.20 & 79.33 & 81.57 & 79.83 \\
        UCF101          & 83.17 & 80.37 & 81.07 & 80.43 & 83.30 & 83.07 & 83.40 & 83.63 \\
        \midrule
        \textit{Average (Seen)} & 81.55 & 79.76 & 80.10 & 79.63 & 82.70 & 82.06 & 82.26 & 81.34 \\
        \midrule
        \textit{Average (H.M)} & 75.94 & 77.33 & 76.79 & 76.74 & 77.42 & 78.32 & 78.19 & 78.15 \\
    \bottomrule[0.1em]
    \end{tabular}}
    \caption{Comparison of accuracy (\%) on unseen classes between Softmax and OVE-PG (Ours) for CoOP, CoCoOp, MaPLe, and APEX.}
    \label{tab:seen}
\end{table*}

\begin{table}[!ht]
\resizebox{1.0\columnwidth}{!}{
\begin{tabular}{l|ccc|ccc|ccc}
\toprule[0.15em]
        & \multicolumn{3}{c|}{\textbf{DTD}}     & \multicolumn{3}{c|}{\textbf{EuroSAT}} & \multicolumn{3}{c}{\textbf{FGVC Aircraft}} \\
        & Seen     & Unseen    & HM   & Seen    & Unseen    & HM    & Seen   & Unseen   & HM   \\ \midrule
SoftMax & 79.37 (2.31) & 54.97 (1.36) & 64.95 (1.71) & 87.77 (1.68) & 55.20 (3.55) & 67.78 (2.28) & 33.60 (4.89) & 22.93 (17.46) & 27.26 (7.64) \\
VPT     & 48.50 (1.30) & 43.77 (2.04) & 46.01 (1.59) & 60.83 (5.11) & 65.90 (3.04) & 63.26 (3.81) & 26.13 (1.93) & 29.30 (0.53) & 27.62 (0.83) \\
OVE-PG  & 77.43 (0.81) & 55.67 (3.25) & 64.77 (1.30) & 85.27 (0.32) & 71.23 (3.45) & 77.62 (0.59) & 32.73 (0.16) & 36.23 (0.86) & 34.39 (0.27) \\ \midrule
        & \multicolumn{3}{c|}{\textbf{Flowers102}} & \multicolumn{3}{c|}{\textbf{Stanford Cars}}    & \multicolumn{3}{c}{\textbf{UCF101}}  \\
        & Seen     & Unseen    & HM   & Seen    & Unseen    & HM    & Seen   & Unseen   & HM   \\ \midrule
SoftMax & 95.27 (0.61) & 72.47 (1.04) & 82.32 (0.77) & 72.30 (1.00) & 72.73 (1.78) & 72.51 (1.28) & 83.17 (1.06) & 72.43 (4.41) & 77.43 (1.71) \\
VPT     & 66.17 (0.64) & 76.40 (0.87) & 70.92 (0.74) & 62.37 (0.49) & 73.60 (0.70) & 67.52 (0.58) & 69.47 (0.47) & 72.60 (2.20) & 71.00 (0.77) \\
OVE-PG  & 92.60 (0.35) & 73.83 (0.81) & 82.16 (0.48) & 67.37 (1.05) & 75.23 (0.57) & 71.08 (0.74) & 80.37 (1.61) & 76.20 (0.32) & 78.23 (0.53) \\
\bottomrule[0.12em]
\end{tabular}
}
\caption{Comparison of accuracy (\%) on seen-to-unseen generalization among Softmax, Bayesian VPT \citep{derakhshani2023bayesian} and OVE-PG (Ours) for CoOp.}
\label{table:coop_all}
\end{table}

\begin{table}[!ht]
\resizebox{1.0\columnwidth}{!}{
\begin{tabular}{l|ccc|ccc|ccc}
\toprule[0.15em]
        & \multicolumn{3}{c|}{\textbf{DTD}}     & \multicolumn{3}{c|}{\textbf{EuroSAT}} & \multicolumn{3}{c}{\textbf{FGVC Aircraft}} \\
        & Seen     & Unseen    & HM   & Seen    & Unseen    & HM    & Seen   & Unseen   & HM   \\ \midrule
SoftMax & 74.30 (2.25) & 55.07 (2.60) & 63.26 (2.41) & 84.83 (0.92) & 66.63 (10.26) & 74.64 (1.69) & 35.83 (0.25) & 34.13 (1.27) & 34.96 (0.42) \\
VPT$^{\dagger}$     & 75.60 (0.79) & 59.17 (0.98) & 66.38 (0.87) & 64.20 (0.46) & 69.10 (0.95) & 66.56 (0.62) & 34.67 (0.21) & 34.80 (0.17) & 34.73 (0.19) \\
OVE-PG  & 75.17 (0.97) & 56.17 (3.22) & 64.30 (1.49) & 84.73 (1.66) & 68.03 (3.99) & 75.47 (2.34) & 35.79 (0.87) & 33.60 (0.46) & 34.66 (0.60) \\
\midrule
        & \multicolumn{3}{c|}{\textbf{Flowers102}} & \multicolumn{3}{c|}{\textbf{Stanford Cars}}    & \multicolumn{3}{c}{\textbf{UCF101}}  \\
        & Seen     & Unseen    & HM   & Seen    & Unseen    & HM    & Seen   & Unseen   & HM   \\ \midrule
SoftMax & 92.53 (0.55) & 74.07 (0.46) & 82.28 (0.50) & 69.80 (0.62) & 74.70 (0.98) & 72.17 (0.76) & 81.07 (0.59) & 72.63 (1.08) & 76.62 (0.76) \\
VPT$^{\dagger}$     & 93.20 (0.30) & 73.77 (1.47) & 82.35 (0.50) & 69.00 (0.46) & 75.30 (0.26) & 72.01 (0.33) & 80.13 (0.21) & 78.17 (1.03) & 79.14 (0.35) \\
OVE-PG  & 92.20 (0.71) & 74.33 (0.93) & 82.31 (0.81) & 69.83 (0.61) & 75.50 (0.36) & 72.55 (0.45) & 80.43 (0.86) & 73.27 (0.96) & 76.68 (0.91) \\
\bottomrule[0.12em]
\end{tabular}
}
\caption{Comparison of accuracy (\%) on seen-to-unseen generalization among Softmax, Bayesian VPT \citep{derakhshani2023bayesian} and OVE-PG (Ours) for CoCoOp. $^{\dagger}$ indicates that we conducted the experiments based on our implementation for CoCoOp, as the original work was only conducted on CoOp.}
\label{table:cocoop_all}
\end{table}

\begin{table}[t]
\centering
\resizebox{1.0\columnwidth}{!}{
\begin{tabular}{cc|cc|cc|cc|cc|cc}
\toprule[0.15em]
\multicolumn{6}{c}{\textbf{CoOp}} & \multicolumn{6}{c}{\textbf{CoCoOp}} \\
\midrule
\multicolumn{2}{c|}{SoftMax} & \multicolumn{2}{c|}{VPT} & \multicolumn{2}{c|}{OVE-PG} & \multicolumn{2}{c|}{SoftMax} & \multicolumn{2}{c|}{VPT} & \multicolumn{2}{c}{OVE-PG} \\
AHM        & \# Param (K)       & AHM      & \# Param (K)     & AHM        & \# Param (K)       & AHM        & \# Param (K)       & AHM      & \# Param (K)      & AHM        & \# Param (K)       \\
\midrule
65.38 & 8.19 & 57.72 & 3.07 & \textbf{68.04} & 8.19 & 67.32 & 35.36 & 66.86 & 659.46 & 67.66 & 35.36 \\
\bottomrule[0.12em]
\end{tabular}}
\caption{Comparison of the averaged harmonic mean (AHM) across six different datasets and the number of learnable parameters (\# Param). Higher AHM and fewer parameters are better. The best values are highlighted in \textbf{bold}.}
\label{tab:overall}
\end{table}

\begin{table}[t]
\centering
\resizebox{1.0\columnwidth}{!}{
\begin{tabular}{l|l|c|cccccccccc|c}
\toprule[0.15em]
 & & \textbf{Source} & \multicolumn{10}{|c|}{\textbf{Target}} & \multirow{2}{*}{\textbf{AVG.}}  \\ 
 & & ImageNet & Caltech101 & DTD & EuroSAT & Aircraft & Food101 & Flowers102 & Pets & Cars & SUN397 & UCF101 &    \\ \midrule
\multirow{2}{*}{CoOp} & SoftMax & 76.5 & 95.9 & 51.7 & 51.3 & 26.7 & 90.0 & 72.6 & 90.7  & 62.8 & 71.8 & 67.2 & 68.84  \\
 & OVE-PG & 74.5 & 95.9 & 57.4 & 61.7 & 23.8 & 89.3 & 71.6 & 89.8 & 63.6 & 71.0 & 69.9 & 69.86   \\
 \midrule
\multirow{2}{*}{CoCoOp} & SoftMax & 75.9 & 95.7 & 54.5 & 52.5 & 24.2 & 89.7 & 73.8 & 90.7 & 65.2 & 73.2 & 69.2 & 69.51  \\
 & OVE-PG & 76.1 & 96.2 & 55.4 & 55.6 & 26.8 & 89.3 & 71.5 & 90.6 & 64.9 & 71.5 & 70.6 & 69.86  \\
\bottomrule[0.12em]
\end{tabular}}
\caption{Extended results for \autoref{table:cross_eval}. We further provide the evaluation results on Caltech101, Food101, Oxford Pets, SUN397 datasets.}
\label{tab:extended_cross_eval}
\end{table}

\begin{table}[!ht]
\centering
\resizebox{1.0\columnwidth}{!}{
\begin{tabular}{l|ccccccc|ccccccc}
\toprule[0.15em]
& \multicolumn{7}{c|}{\textbf{DTD}} & \multicolumn{7}{c}{\textbf{EuroSAT}} \\
\midrule
$\beta$ & 0.1 & 0.2 & 0.3 & 0.4 & 0.5 & 0.6 & 0.7 & 0.1 & 0.2 & 0.3 & 0.4 & 0.5 & 0.6 & 0.7 \\ \midrule
\textbf{Seen} & 73.8 & 77.4 & 74.8 & 76.3 & 75.8 & 76.9 & 75.1 & 85.1 & 85.3 & 84.1 & 83.0 & 83.7 & 78.8 & 80.0 \\
\textbf{Unseen} & 54.2 & 58.6 & 53.4 & 56.3 & 55.2 & 55.7 & 52.9 & 70.4 & 72.7 & 73.9 & 71.6 & 74.7 & 69.1 & 70.8 \\
\textbf{HM} & 62.50 & 66.70 & 62.31 & 64.79 & 63.88 & 64.61 & 62.07 & 77.06 & 78.50 & 78.67 & 76.88 & 78.94 & 73.63 & 75.12 \\ \midrule
& \multicolumn{7}{c|}{\textbf{FGVC Aircraft}} & \multicolumn{7}{c}{\textbf{Flowers102}} \\
\midrule
$\beta$ & 0.1 & 0.2 & 0.3 & 0.4 & 0.5 & 0.6 & 0.7 & 0.1 & 0.2 & 0.3 & 0.4 & 0.5 & 0.6 & 0.7 \\ \midrule
\textbf{Seen} & 32.0 & 32.7 & 32.8 & 32.4 & 33.6 & 31.5 & 32.7 & 91.5 & 91.2 & 91.3 & 92.7 & 90.0 & 90.7 & 87.8 \\
\textbf{Unseen} & 28.1 & 36.2 & 34.6 & 34.6 & 34.8 & 33.5 & 36.2 & 75.0 & 73.8 & 75.2 & 75.6 & 75.5 & 75.7 & 76.6 \\
\textbf{HM} & 29.92 & 34.36 & 33.68 & 33.46 & 34.19 & 32.47 & 34.36 & 82.43 & 81.58 & 82.47 & 83.28 & 82.11 & 82.52 & 81.82 \\ \midrule
& \multicolumn{7}{c|}{\textbf{Stanford Cars}} & \multicolumn{7}{c}{\textbf{EuroSAT}} \\
\midrule
$\beta$ & 0.1 & 0.2 & 0.3 & 0.4 & 0.5 & 0.6 & 0.7 & 0.1 & 0.2 & 0.3 & 0.4 & 0.5 & 0.6 & 0.7 \\ \midrule
\textbf{Seen} & 67.0 & 67.4 & 67.2 & 68.0 & 67.8 & 67.5 & 67.1 & 79.6 & 80.4 & 80.5 & 79.8 & 78.7 & 79.5 & 79.0 \\
\textbf{Unseen} & 74.9 & 76.6 & 74.2 & 75.0 & 75.6 & 75.3 & 75.6 & 78.3 & 78.7 & 78.9 & 78.7 & 77.4 & 77.8 & 78.6 \\
\textbf{HM} & 29.92 & 34.36 & 33.68 & 33.46 & 34.19 & 32.47 & 34.36 & 82.43 & 81.58 & 82.47 & 83.28 & 82.11 & 82.52 & 81.82 \\
\bottomrule[0.12em]
\end{tabular}}
\caption{Extended results for \autoref{fig:sensitivity}. We further provide the results for sensitivity analyses on unseen category and harmonic mean (HM) of seen and unseen categories.}
\label{tab:sensitivity}
\end{table}

\paragraph{Extended Results for \autoref{tab:unseen}.} Here, we provide the performance comparison for seen classes and harmonic mean between seen and unseen categories in \autoref{tab:seen}. While OVE-PG introduces slight performance degradation in seen classes, it achieves higher or comparable performance from the perspective of the harmonic mean. Also, \autoref{table:coop_all} and \autoref{table:cocoop_all} demonstrates the extended results for \autoref{tab:unseen} on six datasets, showing classification accuracy on seen and unseen categories and their harmonic mean, respectively. We also report the averaged harmonic mean across six datasets and the number of learnable parameters for each method in \autoref{tab:overall}. Notably, OVE-PG on CoOp achieves the best average performance in terms of harmonic mean among the six approaches (i.e., SoftMax, Bayesian VPT, and OVE-PG for both CoOp and CoCoOp), indicating that our OVE-PG effectively regularizes the model for fitting both seen and unseen categories. While our method requires the smallest training cost, except for vanilla CoOp, it can significantly improve the overall performance of prompt learning in VLMs.

\paragraph{Extended Results for \autoref{table:cross_eval}.} We also provide the extended results for \autoref{table:cross_eval} in \autoref{tab:extended_cross_eval}. We evaluate the trained models on additional evaluation datasets including Caltech101 \citep{fei2004learning}, Food101 \citep{bossard2014food}, Oxford Pets \citep{parkhi2012cats}, SUN397 \citep{xiao2010sun}. For both CoOp and CoCoOp, our proposed method showed competitive performance on these additional larger datasets consistently outperforms the SoftMax baseline on average.

\paragraph{Extended Results for  and \autoref{fig:sensitivity}.} We further report the numerical values for \autoref{fig:sensitivity} in \autoref{tab:sensitivity}. The values correspond to the results using a seed number of 1. Also, we further provide the results across different $\beta$ both seen and unseen, and their harmonic mean, unlike \autoref{fig:sensitivity} in main manuscript which is restricted from the number of page limitation. In overall, we can conclude that similar observation also holds for seen and harmonic mean as well as for unseen in \autoref{fig:sensitivity} that within the midle range of $\beta$ (i.e., 0.2 -- 0.5), OVE-PG shows robustness regardless of the specific choice of $\beta$.
\vfill
\end{document}